\documentclass[default]{sn-jnl}

\usepackage{lineno,hyperref}
\usepackage{times}
\usepackage{url}
\usepackage{latexsym}
\usepackage{xcolor} 
\usepackage{soul}
\usepackage{microtype, float}
\usepackage{graphicx}
\usepackage{booktabs} 
\usepackage{microtype}
\usepackage{latexsym}
\usepackage{float}
\usepackage{multirow}
\usepackage{enumitem}
\usepackage{tabularx}
\usepackage{arydshln}
\usepackage{booktabs,fixltx2e}
\usepackage{amssymb,mathtools}
\usepackage{eqnarray} 
\usepackage{hyperref}
\usepackage{microtype}
\usepackage{subfigure}
\usepackage{pbox}
\usepackage{amsmath}
\usepackage{pifont}
\modulolinenumbers[5]
\usepackage[normalem]{ulem}
\usepackage[symbol]{footmisc} 

\begin{document}
\title[Interpretable Image Emotion Recognition using Domain Adaptation]{Interpretable Image Emotion Recognition: A Domain Adaptation Approach Using Facial Expressions}

\author[1]{\fnm{Puneet} \sur{Kumar}}\email{puneet.kumar@oulu.fi}
\author*[2]{\fnm{Balasubramanian} \sur{Raman}}\email{bala@cs.iitr.ac.in}

\affil[1]{\orgdiv{Center for Machine Vision and Signal Analysis}, \orgname{University of Oulu, Finland}}
\affil[2]{\orgdiv{Department of Computer Science and Engineering}, \orgname{Indian Institute of Technology Roorkee, India}}

\abstract{This paper proposes a feature-based domain adaptation technique for identifying emotions in generic images, encompassing both facial and non-facial objects, as well as non-human components. This approach addresses the challenge of the limited availability of pre-trained models and well-annotated datasets for Image Emotion Recognition (IER). Initially, a deep-learning-based Facial Expression Recognition (FER) system is developed, classifying facial images into discrete emotion classes. Maintaining the same network architecture, this FER system is then adapted to recognize emotions in generic images through the application of discrepancy loss, enabling the model to effectively learn IER features while classifying emotions into categories such as `happy,' `sad,' `hate,' and `anger.' Additionally, a novel interpretability method, Divide and Conquer based Shap (DnCShap), is introduced to elucidate the visual features most relevant for emotion recognition. The proposed IER system demonstrated emotion classification accuracies of 61.86\% for the IAPSa dataset, 62.47\% for the ArtPhoto dataset, 70.78\% for the FI dataset, and 59.72\% for the EMOTIC dataset. The system effectively identifies the important visual features that lead to specific emotion classifications and also provides detailed embedding plots explaining the predictions, enhancing the understanding and trust in AI-driven emotion recognition systems.}  

\keywords{Interpretable AI, Transfer Learning, Domain Adaptation, Image Emotion Recognition, Discrepancy Loss.}

\maketitle

\section{Introduction}\label{sec:introduction}
Humans predominantly portray emotion-related information visually. The images and videos depicting various emotions may contain facial contents, human activities, different backgrounds, and non-human objects. There is a need to develop computational systems that can identify the emotional information expressed in them. These systems study people's emotional responses to visual information~\cite{joshi2011aesthetics} and are used in animation, gaming, marketing, entertainment, graphics, and lie detection~\cite{kim2018building,joshi2011aesthetics}. The semantic-level Image Emotion Recognition (IER) analysis~\cite{long2015fully} has been explored by the researchers; however, affective level analysis is more difficult \cite{kim2018building, rao2019learning,machajdik2010affective}. The recent progress of deep learning has caused an enormous performance boost for image recognition and object classification~\cite{krizhevsky2012imagenet, long2015fully}. Deep learning-based approaches have been successfully used for Facial Expression Recognition (FER); however, emotion analysis in generic images is a very complex task as they may include human, non-facial components, and non-human objects \cite{kumar2023affective}. The emotional expression in images is sometimes attributed to high-level visual features such as background, and facial structure, whereas, sometimes, low-level image features such as texture, edge, color, and shape determine the emotional portrayal in images.\vspace{.02in}

In the context of emotion recognition in the visual domain, while Facial Expression Recognition (FER) has been extensively studied \cite{li2018deep, minaee2021deep, schroff2015facenet}, its methodologies cannot be directly transferred to Image Emotion Recognition (IER) for generic images without adaptation. The absence of sufficient pre-trained models and well-annotated datasets for IER compared to FER presents significant challenges. To bridge this gap, our proposed system employs a domain adaptation strategy, a special case of transfer learning where the source and target domains share identical feature spaces but have different data distributions. This approach utilizes the advances in FER to enhance IER by adapting successful FER methodologies to broader emotional contexts in images, ensuring the system effectively learns IER features while maintaining robustness in FER applications. By maintaining the same architecture for both FER and IER and employing discrepancy loss, we align the target domain's distribution without altering the network architecture, allowing for seamless adaptation across both domains \cite{nguyen2024deep, rozantsev2018residual, pan2009survey}. This feature-based strategy enhances the model’s ability to generalize across facial and generic images, which is inspired by the understanding that emotions, whether captured through facial expressions or generic images, share underlying affective properties that can be modeled similarly.

The proposed system classifies an image into discrete categories using an adapted FER model, constructed using VGG16, residual blocks, convolution, max pooling, and dense layers, trained with the Adam optimizer. This main FER model and the adapted IER model are trained simultaneously. The discrepancy loss minimizes differences between the models, enabling the adapted IER model to effectively learn the IER domain's distribution. The DnCShap interpretability approach, detailed later in Section~\ref{sec:dncshap}, enhances this model by identifying the key features contributing most towards recognizing specific emotion classes. {While deep learning methods have achieved good performance in IER, they often function as ``black-box'' solutions with limited transparency \cite{kumar2024measuring}. In the context of IER, interpretability is particularly crucial because emotional cues can be subjective and multifaceted, making it essential for practitioners and end-users to understand how decisions are being made. This lack of interpretability poses challenges in real-world applications where accountability and user trust are paramount \cite{kumar2024synthesizing,broniatowski2021psychological}. The proposed DnCShap interpretability approach addresses these concerns by highlighting the salient features that most strongly influence model predictions, thereby bridging the gap between high accuracy and user-centric transparency in IER systems. By revealing the internal decision-making pathways, DnCShap ensures that stakeholders can comprehend the rationale behind specific emotion classifications, ultimately enhancing trust and facilitating broader adoption in sensitive or safety-critical domains.}

{The proposed domain adaptation approach minimizes discrepancy loss, enabling an effective transfer of learned representations from FER to broader IER task. This is achieved by systematically aligning source (facial) and target (generic) distributions and constraining the output probabilities of these two tasks to be consistent through the use of a discrepancy loss. Compared to existing methods, which typically tackle face-only or highly constrained image data, the proposed system preserves robust features pertinent to emotion classification. Furthermore, the newly introduced DnCShap interpretability module provides explanations at both feature and layer-wise levels, detailing what parts of an image drive the model’s decisions—an aspect rarely addressed in prior domain adaptation setups. By uniting robust domain adaptation with detailed interpretability, this results in an interpretable emotion recognition system, advancing the state-of-the-art in IER~\cite{long2015fully, rao2019learning}.} 

The experimental results on the International Affective Picture System Subset a (IAPSa), ArtPhoto, Flicker \& Instagram, and The Emotions In Context (EMOTIC) datasets demonstrate emotion recognition accuracies of 61.86\%, 62.47\%, 70.78\% and 59.72\% respectively, which significantly exceed those reported in previous studies, validating the effectiveness of our approach. The architecture of the proposed IER system and the corresponding domain adaptation strategy have been refined through these studies, highlighting the critical link between the identified gaps in existing models and our contributions to addressing these challenges. The key contributions of this paper are listed as follows.\vspace{.03in}
\begin{enumerate}
\item A novel IER system based on domain adaptation techniques has been proposed to classify generic images containing facial, non-facial, and non-human components into discrete emotion classes.\vspace{.02in}
\item A domain adaptation strategy for simultaneous training of FER and IER models using discrepancy loss has been introduced. This strategy enables the FER models to adapt for IER and to learn the distribution characteristics of IER datasets effectively.\vspace{.02in}
\item The DnCShap interpretability technique has been incorporated into the IER framework, which identifies the input features that most significantly contribute to the recognition of specific emotional classes, enhancing the model’s transparency and interpretability.\vspace{.02in}
\item A novel strategy for interpreting the layer-by-layer learning process and predictions of the IER system has been developed, providing detailed insights into the neural network's decision-making pathways.\vspace{.02in}
\item The proposed system has been validated across diverse datasets, demonstrating superior accuracy and efficiency. Comparative analysis shows significant advancements over existing methods.\vspace{.03in}
\end{enumerate}

The rest of the manuscript has been organized as follows. Section~\ref{sec:lr} provides an in-depth review of the existing literature and related research in the field. Section~\ref{sec:proposed} delineates the architecture and methodology of the proposed system, elaborating on the innovative features and domain adaptation techniques employed. Section~\ref{sec:implementation} covers the implementation details and presents the experimental results, demonstrating the proposed model's effectiveness. Section~\ref{sec:conclusion} concludes the manuscript, summarizing the key findings, discussing their implications, and suggesting future research directions to advance emotion recognition systems.

\section{Related Works}\label{sec:lr}
This section reviews the state-of-the-art in FER and IER along with domain adaptation and interpretability for IER.

\subsection{Image Emotion Recognition}
Facial expressions are essential for machines to understand human emotions. The domain of FER is robust, encompassing a variety of techniques including micro-expression analysis, face localization, landmark points' analysis, face registration, shape feature analysis, face segmentation, and eye gaze prediction \cite{li2018deep, corneanu2016survey}. Contrarily, IER has started gaining attention more recently. Pioneering work by Kim et al.~\cite{kim2018building} developed a neural-based system that synergizes diverse emotional features from images. Additionally, Rao et al.~\cite{rao2019learning} extended these concepts into hierarchical emotion recognition frameworks, allowing for a layered understanding of emotions. Recent explorations by Kumar et al. \cite{kumar2021hybrid}, Khare et al. \cite{khare2023emotion}, and Cirneanu et al. \cite{cirneanu2023new} further expand the scope of IER by integrating innovative methodologies that address both facial and non-facial components. These advancements align closely with the objectives of our research, which introduces an IER system based on domain adaptation designed to classify complex emotional content in images including facial, non-facial, and non-human elements. Our approach utilizes a domain adaptation strategy using discrepancy loss to enhance the FER model's adaptability for the IER task.

\subsubsection{Facial Expression Recognition}
The FER has significantly evolved from traditional manual feature extraction methods to advanced deep learning approaches \cite{happy2014automatic, minaee2021deep, bisogni2024poser}. Traditional FER involves a multi-step process starting with pre-processing, where techniques like histogram equalization are commonly applied to normalize lighting across images \cite{happy2014automatic}. This is followed by feature extraction using methods such as local binary patterns, which capture textural information \cite{chen2014facial}, and Gabor wavelets, noted for their edge detection capabilities \cite{bartlett2005recognizing}. These traditional techniques, however, often struggle with the variability of facial expressions within the same category. Transitioning to deep learning, these methods have revolutionized FER by enhancing the ability to handle intra-class variations effectively, without the need for manual intervention \cite{verma2018expertnet}. Khorrami et al. demonstrated this with their work on a deep network with zero bias, which showed improved performance on the Cohn-Kanade (CK) dataset \cite{khorrami2015deep}. Further advancements include Aneja et al.'s deep FER model that facilitates animated and human facial expression mapping \cite{aneja2016modeling}, highlighting the versatility of deep learning models.

Moreover, the field has seen innovations such as Castellano et al.’s real-time automatic FER system, which is particularly useful in dynamic environments such as during the COVID-19 pandemic \cite{castellano2023automatic}. Malik et al. focused on enhancing FER in low-resolution images, crucial for surveillance contexts \cite{malik2021towards}, and Chowdary et al. have worked on improving the nuances of interaction dynamics within human-computer interfaces \cite{chowdary2023deep}. Additionally, Jain et al. have developed a hyperparameter-tuned deep learning model for autonomous vehicles, ensuring driver safety through effective emotion recognition \cite{jain2023automated}. Vignesh et al. introduced a novel model using segmentation and VGG-19 architecture to achieve high accuracy in emotion detection \cite{vignesh2023novel}, and Talaat has implemented a real-time system for children with autism, combining deep learning with IoT technologies for behavioural therapy \cite{talaat2023real}.

These technological enhancements not only demonstrate significant strides in accuracy and real-time processing but also point to a gap in the understanding of the internal decision-making processes of FER models, as discussed by Pedro et al. and Singh et al. \cite{fernandez2019feratt, singh2018hierarchical}. Our research aims to bridge this gap by enhancing the explainability and transparency of FER systems, making them more accessible and comprehensible to users.

\subsubsection{Generic Image Emotion Recognition}
The IER has evolved from relying solely on low-level features such as shape, colour, and edge \cite{hanjalic2006extracting} to incorporating mid-level features like composition and optical balance, which enhance the aesthetic interpretation of images \cite{joshi2011aesthetics}. Pioneering works by Zhao et al. \cite{zhao2014exploring} and Machajdik et al. \cite{machajdik2010affective} have furthered the use of mid-level visual features and semantic visual content, respectively, to deepen emotional analysis. While IER methodologies are often divided into Dimensional Emotion Space (DES) approaches that analyze emotional states through arousal and valence \cite{kim2018building,zhao2016continuous}, and Categorical Emotion States (CES) approaches that classify emotions into discrete classes like happiness, sadness, anger, and hate \cite{rao2019learning, zhao2018discrete, zhao2014exploring}, our research enhances the CES framework by identifying complex emotions in images containing a mix of facial, non-facial, and non-human components.

Traditional IER approaches generally utilize a feature-based semantic analysis, often limited by the manual crafting of low-level features, which fails to capture the full semantic potential of images \cite{hanjalic2006extracting}. Although advances by Zhao et al. \cite{zhao2014exploring} and Machajdik et al. \cite{machajdik2010affective} have expanded the use of mid-level features, there remains a gap in fully integrating all semantic layers effectively. Our work addresses these challenges by implementing a domain adaptation strategy that significantly improves the system's ability to interpret a broad spectrum of visual features cohesively and accurately.

Moreover, the advent of deep learning has revolutionized IER by enabling the extraction and analysis of features without manual intervention through end-to-end methodologies \cite{you2016building, zhao2014exploring, rao2016multi}. These approaches, utilizing Convolutional Neural Network (CNN)-based transfer learning, adapt pre-trained models to new IER tasks efficiently. However, they often struggle with the precise extraction and utilization of both low and mid-level features essential for accurate emotion recognition \cite{rao2019learning}, compounded by the scarcity of large-scale, well-annotated IER datasets and the inherent subjectivity of emotion perception. Our system overcomes these hurdles by incorporating a novel discrepancy loss function within our domain adaptation framework, ensuring that both FER and IER models effectively learn from and enhance each other, improving their performance across diverse datasets and complex emotional scenarios.

\subsection{Domain Adaptation for Image Emotion Recognition}
Domain Adaptation (DA), a specialized form of Transfer Learning, is crucial for enhancing the versatility and accuracy of emotion recognition systems across diverse domains. It specifically addresses the challenge of applying models trained on the source domain to effectively perform on the target domains. Advanced techniques such as deep cross-domain transfer facilitate emotion recognition via joint learning \cite{nguyen2024deep}, and multisource marginal distribution adaptation has been particularly effective in cross-subject and cross-session EEG emotion recognition \cite{chen2021ms}. Additionally, multi-view domain adaptive representation learning has shown promise in EEG-based emotion recognition \cite{li2024multi}. Strategies such as spectral \cite{yang2024spectral}, multi-domain \cite{li2024ms}, and adversarial domain adaptation \cite{liu2024capsnet, cai2023exploiting} push the boundaries of adapting emotion recognition models to new challenges.

Historically, emotion recognition has relied heavily on facial images \cite{minaee2021deep}, with systems often utilizing attention models to enhance FER \cite{fernandez2019feratt}. Domain adaptation in this field encompasses various approaches including `feature-based adaptation' where features are transformed or augmented to bridge the gap between source and target domains \cite{saenko2010adapting, li2013learning}, `instance-based adaptation' which focuses on selecting relevant training samples from the source domain \cite{shekhar2013generalized}, and `model-based adaptation' strategies that involve modifying the model architecture or training strategy, such as using adversarial techniques or discrepancy loss \cite{rozantsev2018residual}.

{Recent works have extended domain adaptation into cross-subject and multi-modal directions specifically targeting emotion recognition. Lin et al.~\cite{lin2020multi} introduced a multi-source domain adaptation framework for visual sentiment classification, demonstrating how fusing knowledge across multiple labeled domains can significantly enhance transferability. Zhao et al.~\cite{zhao2021plug} proposed a ``plug-and-play'' approach to domain adaptation for EEG-based emotion recognition, highlighting the modular reusability of adaptation blocks across different subjects. Similarly, Ahn et al.~\cite{ahn2021cross} investigated cross-corpus speech emotion recognition through few-shot learning and domain adaptation, underscoring the interplay between minimal supervision and acoustic domain shifts. Meanwhile, Zheng et al.~\cite{zheng2025fuzzy} tackled source-free domain adaptation challenges in visual emotion recognition with a fuzzy-aware loss, effectively mitigating noisy pseudo-label issues in the target domain. Collectively, these advances reinforce the necessity of robust, noise-tolerant adaptation methods in diverse affective computing scenarios.}

We employ a feature-based domain adaptation approach, focusing directly on visual modalities to preserve the integrity of emotional data, thereby avoiding the information loss associated with methods like those using image captions for domain adaptation \cite{kumar2021domain}. This approach is exemplified by the FaceNet architecture, which learns mappings from facial images into representational spaces based on face-similarity measures \cite{schroff2015facenet}. However, these FER models often falter when applied to non-facial images, necessitating adaptations for broader IER applications. Our system adapts pre-trained FER models to effectively recognize emotions in generic non-facial images using this feature-based domain adaptation strategy. Employing discrepancy loss to align the feature distributions of FER and IER models without altering the network architecture ensures optimal transfer and adaptation of learned features across domains \cite{rozantsev2018residual}. This method addresses the challenge of varied emotional cues in different image contexts and is capable of handling diverse datasets. It not only enhances the accuracy and applicability of our system but also interprets a wide range of visual features, significantly improving the robustness of emotion recognition.

\subsection{Interpretability and Explainability}
Interpretability and Explainability are increasingly vital in the field of AI and machine learning, particularly within the domain of emotion recognition \cite{kumar2024measuring,kumar2024synthesizing}. These concepts are essential for building trust in AI systems, as they allow users to understand and validate the decision-making processes of the models \cite{broniatowski2021psychological}. Interpretable emotion analysis has become a significant trend, with research expanding across both visual \cite{song2024emotional, paskaleva2024unified} and other modalities \cite{li2024explanation, kumar2023interpretable}, underscoring the need for systems that users can understand and interact with confidently.

The process of making a deep learning-based classifier’s operations transparent involves elucidating the mechanisms that lead to specific outputs. Ribeiro et al. \cite{ribeiro2016should} introduced an influential approach by developing an algorithm that explains predictions by approximating a classifier locally with an interpretable model and identifying the parts of the input most responsible for the output. Similarly, Shrikumar et al. \cite{shrikumar2017learning} presented a technique to decompose the predictions of a neural network by backpropagating the contributions of each neuron to understand their impact on the final decision. These methodologies, while groundbreaking, often fall short of revealing how the neural network weights are trained and adjusted during the learning process, a gap that remains largely unaddressed in current literature.

{With the increasing complexity of emotion recognition systems, the need for clarity in how decisions are made is crucial. Recent efforts by Kang et al.~\cite{kang2022privacy} have introduced a privacy-preserving adversarial domain adaptation approach, which not only protects sensitive information but also clarifies feature group processing, linking domain adaptation with clear, actionable insights. Asokan et al.~\cite{asokan2022interpretability} utilize concept activation vectors to illuminate the contributions of multimodal emotion cues, like facial expressions and vocal tones, to decision-making processes within models. Furthermore, Malik et al.~\cite{malik2021towards} have enhanced the transparency of facial emotion recognition systems by localizing and accentuating critical facial regions, significantly boosting user trust and understanding of automated affective assessments.}

Our research builds on these foundational works by proposing a system that not only explains emotion recognition outcomes but also provides insights into the training dynamics of network weights. This is crucial for the accurate interpretation of subtle emotional cues, which is essential in effective human-computer interaction. By enhancing transparency and interpretability, we aim to create more reliable AI tools for emotion analysis \cite{kumar2024interpretable}. Our findings show that integrating real-time data interpretation and dynamic adjustment of network weights not only meets but exceeds theoretical expectations and outperforms traditional models in initial tests. These results underline the practical benefits of our approach in effectively addressing gaps in existing models. 

\section{Methodology}\label{sec:proposed}
This section proposes a novel approach for transferring the FER model to the IER task using a feature-based domain adaptation strategy. We maintain the same architecture for both FER and IER and employ discrepancy loss to effectively align the target domain's distribution, thereby enhancing the model's capability to learn IER features. 

\subsection{Problem Formulation}\label{sec:prob}\vspace{-.03in}
The target task $\mathbb{T}_i$ is defined as IER for feature space $\mathbb{X'}$ and target domain $\mathbb{D}_i$, where $\mathbb{D}_i \subset \mathbb{X'}$. The source domain $\mathbb{D}_f$ and feature space $\mathbb{X}$, where $\mathbb{D}_f \subset \mathbb{X}$, are designated for FER, denoted as $\mathbb{T}_f$. The task $\mathbb{T}_i$ in $\mathbb{D}_i$ utilizes the information from $\mathbb{T}_f$ and $\mathbb{D}_f$. Given the different distribution of their data points, i.e., images, $\mathbb{D}_i$ undergoes a statistical adaptation to align with the source domain's feature space $\mathbb{X}$. According to Pan et al. \cite{pan2009survey}, this adaptation process involving probabilistic transformations is consistent with Domain Adaptation as both $\mathbb{D}_f$ and $\mathbb{D}_i$ perform the same task, share the same feature space, but have different marginal distributions of data points.

\subsection{Facial Expression Recognition}\label{sec5.1.2}

\begin{figure}[]
\centering
\includegraphics[width=1\textwidth]{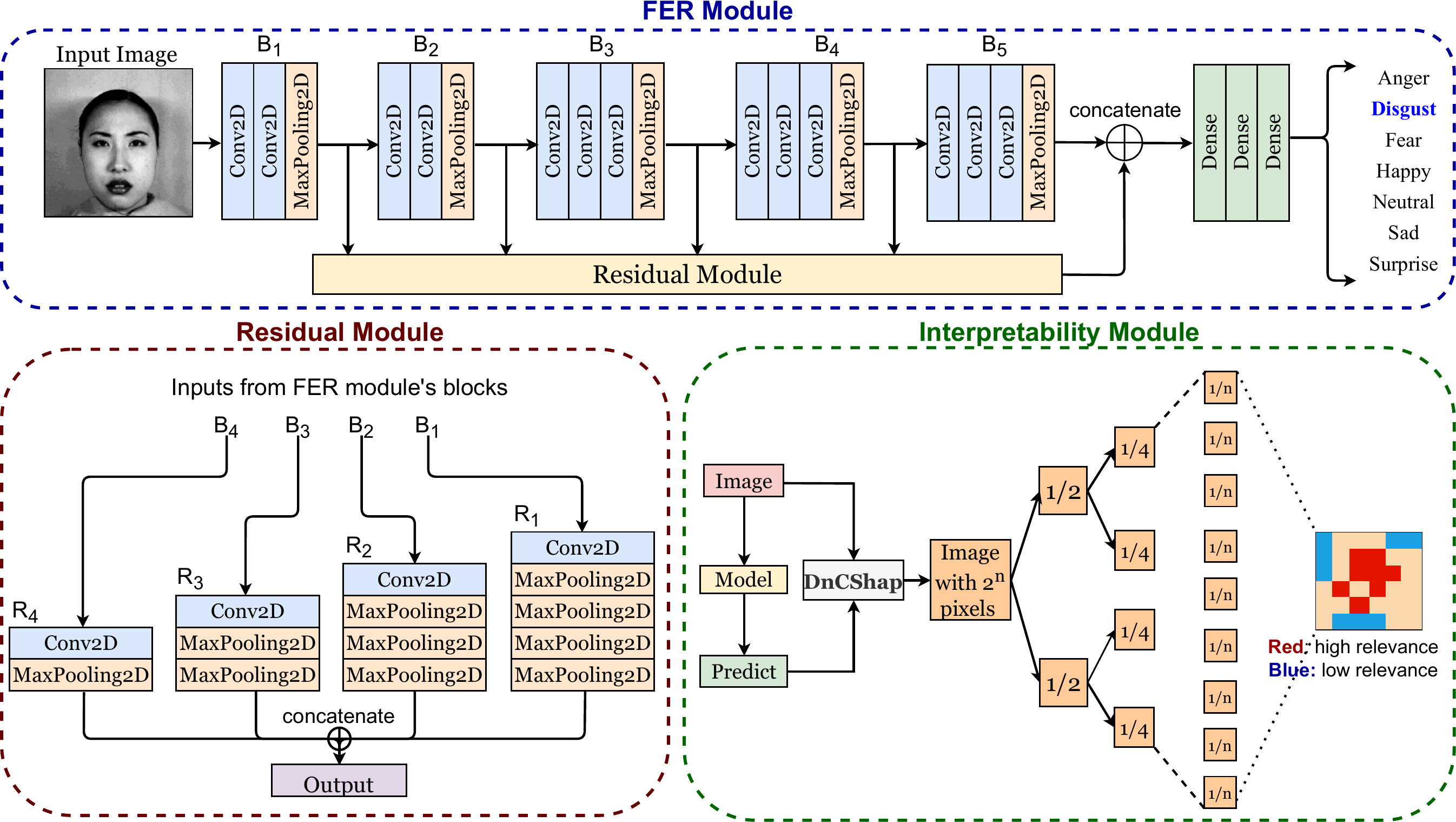}
\caption{The proposed FER system's architecture contains FER, residual, and interpretability modules. The input is processed through FER module's layers B1 to B5, followed by residual module's parallel branches R1 to R4. The interpretability module uses DnCShap to analyze pixel relevance for emotion recognition.} 

\label{fig:archi}
\end{figure}

Fig. \ref{fig:archi} illustrates the architecture of the proposed FER system, which has been constructed on top of a pre-trained VGG16 network \cite{simonyan2014very} by incorporating four residual layers. Its structure comprises five blocks, \(B_k\) where \(k \in \{1, 2, \ldots, 5\}\). Block \(B1\) includes two convolutional layers with 64 channels, \(B2\) has two convolutional layers with $128$ channels, while \(B3\) and \(B4\) contain three convolutional layers with $256$ and $512$ channels, respectively. \(B5\) is configured identically to \(B4\). Each of these blocks employs filters of size \(3 \times 3\) followed by a max-pool layer with a stride of \(2, 2\). 

The residual module has been included to mitigate the issues of exploding and vanishing gradients commonly associated with deep neural networks \cite{he2016deep}. It links four residual layers, \(R_l\) where \(l \in \{1, 2, 3, 4\}\), that process the outputs of blocks \(B1\), \(B2\), \(B3\), and \(B4\) in parallel. \(R1\) comprises four max-pooling layers and one convolutional layer with $512$ channels, \(R2\) includes three max-pooling layers and one convolutional layer with $512$ channels, \(R3\) and \(R4\) incorporate two and one max-pooling layers, respectively, each followed by a single convolutional layer. The outputs from these individual residual layers (\(R1\) to \(R4\)) are then concatenated into a single tensor, which is further concatenated with the output from \(B5\). This concatenated output is then flattened and passed through three dense layers of sizes \((1, 2048)\), \((1, 2048)\), and \((1, 7)\). By concatenating outputs from multiple residual layers, the model integrates and preserves information across processing depths, enhancing feature extraction and generalization capabilities. This configuration allows each network section to learn features at varying levels of abstraction, improving its ability to handle complex tasks like emotion recognition from facial expressions.

The interpretability module employs the Divide and Conquer based SHAP (DnCShap) method to attribute significance to individual pixels in contributing to the model’s decision-making process. By decomposing the input image into its constituent features and assessing their impact, this module provides visual insights into which aspects of the image are most relevant for each recognized emotion. The proposed two-fold interpretability approach for IER, encompassing both feature-wise and layer-wise analyses, is detailed in Section \ref{sec:inter}.

\subsection{Image Emotion Recognition}\label{sec:method1ch5b}
\begin{figure}[]
\centering
\includegraphics[width=1\textwidth]{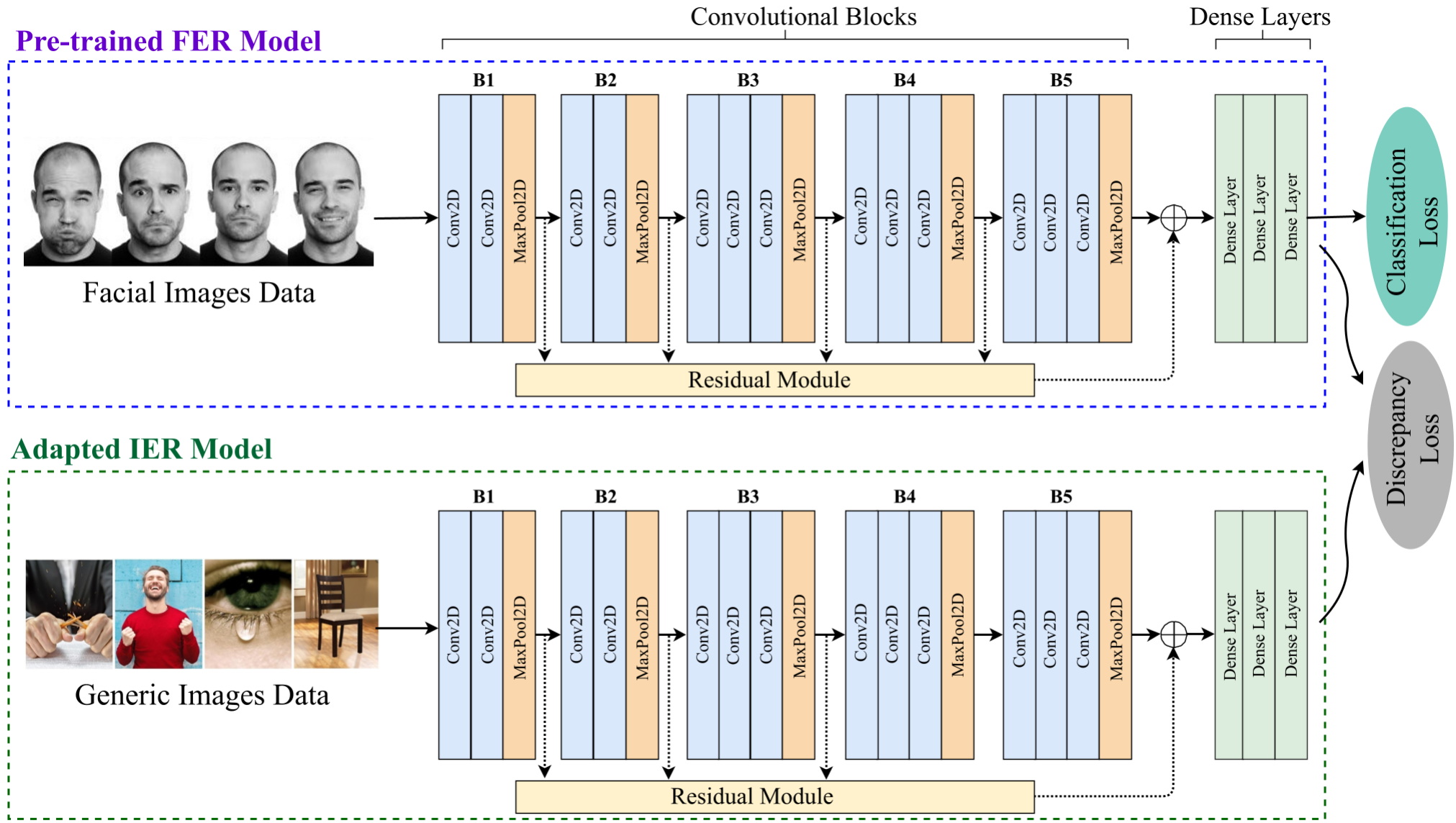} 
\caption{Architecture of the proposed IER system, adapted from the pre-trained FER model. The convolutional blocks B1 to B5 and residual modules facilitate the feature adaptation necessary for handling generic images in the IER model. The integration of discrepancy loss demonstrates how domain adaptation techniques optimize the model for diverse emotional cues. The Classification Loss ensures the accurate categorization of emotional states, enhancing the overall performance and effectiveness of the model.}
\label{fig:archich5b}
\end{figure}

Fig.~\ref{fig:archich5b} illustrates the architecture of the proposed IER system, which was refined through ablation studies. This system is aligned with the FER task as detailed in Section \ref{sec5.1.2}. Both the FER and IER models employ the same VGG16 architecture, which includes additional residual, convolutional, max pooling, and dense layers, ensuring consistency across tasks. The training process begins with the FER model using an Adam optimizer, and the IER model, mirroring the FER model’s architecture, is trained in parallel. Discrepancy loss is minimized between the outputs of both models during training to ensure accurate alignment and effective emotion recognition across different types of images.\vspace{.1in}

\noindent {\textit{Preprocessing}: Prior to training, significant preprocessing was applied to both datasets to ensure consistency and reliability of the input data. Initially, all images were resized to a standard dimension of 224x224 pixels and normalized to have zero mean and unit variance to mitigate any model biases due to variations in image scale and illumination. To address discrepancies in labeling conventions across datasets, a mapping scheme was devised that aligns the emotion labels from different sources into a unified set of categories. Furthermore, given the inherent class imbalance in the datasets, a combination of oversampling the minority classes and undersampling the majority classes was employed to ensure an equitable representation of all classes in the training process. These preprocessing steps are crucial as they significantly influence the model's ability to generalize across diverse inputs and directly impact the performance outcomes discussed in Section \ref{sec:resultsch5b}.}\vspace{.1in}

\noindent \textit{Domain Adaptation Strategy}: We employ a feature-based domain adaptation strategy using the same network architecture for both FER and IER tasks. This approach utilizes discrepancy loss to effectively align the target domain's distribution with that of the source. The discrepancy loss, which is minimized after each training epoch, helps the adapted IER model accurately learn the IER domain's distribution, despite the differing marginal distributions of data points between the FER and IER domains. This alignment is critical for the model's ability to perform consistently across both domains. The overall loss function is detailed in Eq.~\ref{eq:eq0ch5b}.\vspace{-.3in}

{  
\begin{eqnarray}
\label{eq:eq0ch5b}
\begin{split}	
&\mathcal{L}_{\text{overall}} = 
\mathcal{L}_{\text{classifier}} + 
\mathcal{L}_{\text{discrepancy}} + 
\lambda||\omega_{(fc)}||_2^2\\
\end{split}
\end{eqnarray}
} \vspace{-.15in}

where $\mathcal{L}_{overall}$, $\mathcal{L}_{classifier}$, and $\mathcal{L}_{discrepancy}$ are overall loss, cross-entropy loss and discrepancy loss. The third term is the regularization term, where $\omega$ and $\lambda$ are the weights of the fully connected layers and the regularization weight, which are calculated experimentally by observing the variance in the output. The computation of classifier loss and discrepancy loss has been explained as follows. The cross-entropy loss has been used to find the output probability of the main FER model and the adapted IER model. For an image sample $v$, distribution $p_1$ in the FER domain and distribution $p_2$ in the IER domain, classifier loss is calculated using Eq.~\ref{eq:eq1ch5b}. \vspace{-.25in} 

\begin{eqnarray}
\label{eq:eq1ch5b}
\begin{split}	
&\mathcal{L}_{\text{classifier}}\ (p_1, p_2) = 
- \sum_{\forall v}^{} {p_{1}(v)log(p_{2}(v))}\\
\end{split}
\end{eqnarray}\vspace{-.1in} 

where $v$ is an image sample whereas $p_{1}$ and $p_{2}$ denote the actual and predicted distributions of $v$, respectively. Further, the absolute difference between the two classifiers' probabilistic outputs has been utilized as discrepancy loss. For the image samples' distribution for the FER classifier, $p_1$ and the predicted distribution for the IER classifier, $p_2$, the discrepancy loss is calculated as per Eq.~\ref{eq:eq2ch5b}. \vspace{-.25in}

\begin{eqnarray}
\label{eq:eq2ch5b}
\begin{split}
& \mathcal{L}_{\text{discrepancy}}(p_1, p_2) =  
\frac{1}{K} \sum_{k=1}^{k=K}  { \lvert o_{1k} - o_{2k} \rvert }
\end{split}
\end{eqnarray}\vspace{-.1in}

where $o_{1k}$ and $o_{2k}$ denote the probabilistic outputs for the distributions $p_1$ and $p_2$, and $K$ is the number of classes. This discrepancy loss is key to ensuring that the adapted model effectively learns the unique distribution of the IER domain, enhancing its predictive accuracy and generalizability across diverse datasets. This strategy significantly improves the adaptability and effectiveness of the existing FER systems for the broader and more complex task of IER, demonstrating the model's enhanced capability to interpret emotional cues in a wide array of image contexts.\vspace{.1in}

\noindent {\textit{Theoretical Underpinnings of Discrepancy Loss}:
Discrepancy loss plays a critical role in our domain adaptation framework by quantitatively assessing and minimizing the disparity between feature distributions of the source (FER) and target (IER) domains. It measures the divergence between the probability distributions of features extracted from the FER and IER domains, specifically at the last hidden layer of the network. As incorporated in the overall loss function (Eq~\ref{eq:eq0ch5b}), $\mathcal{L}_{\text{discrepancy}}$ is instrumental in aligning these domains. The alignment facilitated by discrepancy loss is not merely a statistical alignment of distributions but a strategic enforcement of feature space homogenization. This process reduces the model’s sensitivity to the domain-specific nuances of emotional expression, which are prevalent when transitioning from FER to IER. By minimizing the discrepancy loss, we effectively guide the model to find a common feature space, where both domains' statistical properties are not just aligned but are functionally indistinguishable in terms of their contribution to the model’s output. This ensures that the learned representations are robust, maintaining high accuracy and consistency across varied emotional datasets and contexts. Moreover, the use of discrepancy loss aids in overcoming overfitting to the domain-specific traits of the training data by promoting domain-invariant feature learning. This is critical for applications in real-world scenarios where the diversity of emotional expressions and the contexts in which they are presented can vary significantly from the conditions seen during training. By ensuring that our model learns to generalize across these domains effectively, we enhance its usability and reliability in practical deployments where domain variability is the norm.\vspace{.1in}}

\noindent {\textit{Impact of Discrepancy Loss Function}: The role of the discrepancy loss in our domain adaptation strategy is multifaceted. It is instrumental in training the model to reduce the divergence in output probabilities between the FER and IER domains for similar emotional inputs, facilitating a smoother transition from FER to IER tasks. Additionally, it allows the model to generalize more effectively across varied image contexts where emotional expressions differ significantly from the training data. This enhanced adaptability is crucial for deploying the model in real-world applications where it must perform reliably across different scenarios and datasets.\vspace{.1in}}

In the proposed domain adaptation strategy, each component is strategically crafted to optimize the model's performance in the new IER domain. The cross-entropy loss ensures that the fundamental task of emotion recognition is maintained by effectively training the model to minimize the difference between the actual and predicted emotion classifications in the FER domain. This forms the baseline upon which the domain adaptation builds. The discrepancy loss, on the other hand, is critical for aligning the model to the new IER domain. It specifically addresses the challenge of different marginal distributions between the FER and IER domains. By minimizing this loss, the model is trained to reduce the divergence in output probabilities between the domains for similar emotional inputs. This not only facilitates a smoother transition of the model from FER to IER tasks but also enhances the model’s ability to generalize across varied image contexts where emotional expressions differ significantly from the training data. 

\subsection{Interpretability}\label{sec:inter} 
A two-fold interpretability technique has been proposed to explain the working of the proposed method. It first visualizes the important features of the input image responsible for emotion classification. Then, implementing the proposed layer-wise explainability technique results in the intersection matrices and cluster distances. 

\subsubsection{Feature-wise Interpretability}\label{sec:dncshap} 
This section introduces DnCShap, which integrates the Divide and Conquer approach into the SHAP method to expedite Shapley values' computation. While the exact Shapley value calculation is NP-hard \cite{SHAPley1953value}, SHAP approximates it in quadratic time. DnCShap further reduces the time complexity to linear by dividing the feature set into manageable subsets, computing Shapley values independently, and aggregating them efficiently.\vspace{.1in}

\noindent \textit{Mathematical Proof of Efficiency in DnCShap}:
The Divide and Conquer method implemented in DnCShap divides the feature set into smaller, manageable subsets, upon which Shapley values are independently computed and then aggregated. This division reduces the computational complexity from quadratic to linear time as follows:

Given a feature set \( F \), it is divided into subsets \( F_1 \) and \( F_2 \). For a feature \( f \) in \( F_1 \), Shapley values are calculated using Eq. \ref{eq:4}.

\begin{equation}
\label{eq:4}
\phi_{F_1}(f) = \sum_{S \subseteq F_1 \setminus \{f\}} \frac{\vert S\vert ! \cdot (\vert F_1\vert - \vert S\vert - 1)!}{\vert F_1\vert !} \left[v(S \cup \{f\}) - v(S)\right]
\end{equation}
\vspace{.05in}

The same process is applied to \( F_2 \) and other subsets. As depicted in Eq. \ref{eq:5}, the final Shapley value for \( f \) is approximated by averaging these individual contributions.\vspace{-.1in}
\begin{equation}
\label{eq:5}
\phi(f) = \frac{1}{2} (\phi_{F_1}(f) + \phi_{F_2}(f))
\end{equation}\vspace{-.15in}

This method ensures that each subset computation can be handled independently, dramatically reducing the computational burden and enabling parallel processing.\vspace{.1in}

\begin{figure}[!h]
\centering
\includegraphics[width=0.45\textwidth]{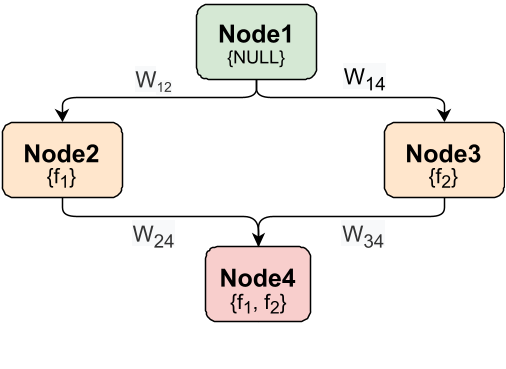} \vspace{-.2in}
\caption{Illustration of Shapley value calculation. Nodes (Node1 to Node4) represent different configurations of feature combinations where Node1 has no features, and subsequent nodes incrementally include features $f_1$ and $f_2$ while $w_{pq}$ denotes the weight coefficient for node pair $Node_p$-$Node_q$.}
\label{fig:SHAP_exp}
\end{figure}

\noindent \textit{Calculating Shapley values}: 
The Shapely values' computation has been demonstrated with an example shown in Fig. \ref{fig:SHAP_exp}. The \textit{Node1} has no feature, \textit{Node4} has two features ($f_1$ and $f_2$) and the rest of the nodes contain one feature each (i.e., $f_1$ and $f_2$, respectively). The marginal contribution of a feature differentiates the predictions of two nodes connected by an edge.  

\noindent The feature $f_1$'s marginal contribution for label $c$ for the model with $f_1$ feature only is computed as per Eq. \ref{eq:eq1ch5a}. \vspace{-.15in} 

{
\begin{equation}	
\label{eq:eq1ch5a}
MC_{f_1,\{f_1\}} = score_{\{f_1\}} - score_{\{\phi\}}\\
\end{equation}
\vspace{.02in}

\noindent where $score_{\{f_1\}}$ is the prediction for label $c$ using the model with feature $f_1$. The overall contribution is computed as per Eq. \ref{eq:eq2ch5a}.\vspace{-.25in}

{
\begin{eqnarray}
\label{eq:eq2ch5a}
\begin{split}	
&SHAP_{\{f_1\}} = w_{12} \times MC_{f_1,\{f_1\}} + w_{34} \times MC_{f_1,\{f_1,f_2\}} \\
\end{split}
\end{eqnarray}
}

\noindent where $w_{12}$ and $w_{34}$ are the coefficient weights. The coefficient weight, $w_i$ is computed using Eq. \ref{eq:eq3ch5a}. \vspace{-.25in}

{
\begin{eqnarray}
\label{eq:eq3ch5a}
\begin{split}	
&w_i = \frac{\lvert S \rvert (\lvert F \rvert - \lvert S \rvert -1)!}{\lvert F \rvert!}\\
\end{split}
\end{eqnarray}
} 

\noindent where $\lvert S \rvert$ denotes no. of features and $|F|$ shows the no. of weights. The values of $w_{12}$ and $w_{34}$ are calculated as per Eq. \ref{eq:eq3ch5a} and Eq. \ref{eq:eq4ch5a} computes the Shapely values. \vspace{-.2in}

{
\begin{eqnarray}
\label{eq:eq4ch5a}
\begin{split}	
&w_{12}=(0!(2-0-1)!)/(2!) = 1/2\\~\vspace{.05in}
&w_{34}=(1!(2-1-1)!)/(2!) = 1/2\\~\vspace{.05in}
&SHAP_{\{f_1\}} = (\frac{1}{2}) \times MC_{f_1,\{f_1\}} + (\frac{1}{2}) \times MC_{f_1,\{f_1,f_2\}}
\end{split}
\end{eqnarray}
}

\noindent \textit{Approximating Shapley values}: 
Instead of re-training the model for a varying number of features, we perturb the input to compute the Shapely values. The incorporation of the Divide and Conquer approach enables the Shapely values' computation in linear time as compared to the exponential time required by the existing methods. It recursively divides the image into halves and computes their Shapley values. The difference between average and predicted values gives the Shapley values of all features. To compute $score1$ and $score2$ denoting the Shapley values of two halves obtained by Divide and Conquer, the property shown in Eq. \ref{eq:eq4} is held true, and $pred\_b$ and $pred\_f$ values are chosen accordingly.

For image $X$ with width $w$ and height $h$, it is divided into two parts $x_1,x_2$ where width is $w/2$ and height is $h$. The two parts are considered as different features, and Eq \ref{eq:eq2ch5a} is used to find their Shapley values. The divide and conquer is continuously applied, and Shapely values are found until the image is divided into a pre-specified (hyperparameter) number of parts. The time complexity is of the order of $T(n)=2T(n/2) + O(1)$, i.e., of the order of $O(n)$, which comes out to be linear time complexity for $n$ number of features.

\subsubsection{Layer-wise Interpretability}
This section proposes a novel technique to explain the predictions of the proposed deep learning-based system. We have defined the term \textit{Intersection Score}, which depicts the correctness of network weights layer by layer. It first visualizes the important features of the input image responsible for emotion classification. The convergence of the emotion clusters is then visualized for various network layers.

\noindent The intersection score is calculated using the following steps.\vspace{.15in}

\noindent \textbf{a)} For each emotion class $i$, calculate the principle components \textbf{($x_i$, $y_i$, $z_i$)} of $p^{th}$ last layer's embedding. For each emotion class $i$ and all three components ($x_i$, $y_i$, $z_i$), mean $m_i$ and standard deviation $\sigma_{\bar{m_i}}$ are calculated using Eq.~\ref{eq:eq2}, where $n_i$ is the number of data points for $i^{th}$ emotion-class and $k$ is the $k^{th}$ data point.\vspace{-.25in}

{\fontsize{10}{12}\selectfont 
\begin{eqnarray}
\label{eq:eq2}
\begin{split}	
\bar{m_i} = \frac{1}{n_i} \sum_{k=1}^{k=n_i} {m_k};\hspace{1mm}
\sigma_{\bar{m_i}} = \frac{1}{n_i} \sum_{k=1}^{k=n_i}({m_k-\bar{m_i}})^2\hspace{2mm}\\
\forall m \in \{x,y,z\}
\end{split}
\end{eqnarray}
}\vspace{-.1in} 

To get the principle components of a layer’s embeddings, first, the embedding vector is extracted and flattened. Then the principal component analysis (PCA) procedure is applied, which involves computing eigenvectors; sorting them by decreasing eigenvalues and choosing k eigenvectors with the largest eigenvalues and using this d $\times$ k eigenvector matrix for transformation where d dimensions.\\

\noindent\textbf{b)} For each emotion class $i$, the range for every principle component is defined using the above parameters with an assumption that most of the values will be within two times the standard deviations of the mean. Range for $i^{th}$ emotion is defined as $[L_i(m),R_i(m)]$ where $L_i(m)$ and $R_i(m)$ are the left and right extreme points of $m^{th}$ component's spread for emotion class $i$. The extreme points are defined in Eq.~\ref{eq:eq3}. The choice of using the double of the standard deviation value to define the range is done under the assumption that the distribution has a symmetric tail on both sides. \vspace{-.2in}

{\fontsize{10}{12}\selectfont 
\begin{eqnarray}
\label{eq:eq3}
\begin{split}	
\hspace{2mm}L_i(m)=\bar{m}_i-2\sigma_{\bar{m_i}}\hspace{1mm}
R_i(m)=\bar{m}_i+2\sigma_{\bar{m_i}}\hspace{1mm} \\
\forall m \in \{x,y,z\} 
\end{split}
\end{eqnarray}
} 

\noindent \textbf{c)} The intersection between emotion classes $i$ and $j$ is calculated as per Eq.~\ref{eq:eq4} and denoted as $I_{i,j}(m)$. Here, four emotion classes, i.e., angry, happy, sad, and neutral, are considered. \vspace{-.2in}

{\fontsize{9.5}{12}\selectfont 
\begin{eqnarray}
\label{eq:eq4}
\begin{split}	
I_{i,j}(m)= \frac{max\{(min(R_i(m),R_j(m))-max(L_i(m),L_j(m))),0\}}
{max(R_i(m),R_j(m))-min(L_i(m),L_j(m))}\\
\forall m \in \{x,y,z\},\hspace{3mm} \forall i,j \in \{1,2,3,4\}
\end{split}
\end{eqnarray}
}\vspace{-.1in}

\noindent Here, $I_{i,j}(m)$ denotes the 
intersection between the spread of $m^{th}$ component's data for emotion classes $i$ and $j$.\\ 

\noindent\textbf{d)} The total intersection between two emotion classes $i$ and $j$ is denoted as $I_{i,j}$. As shown in Eq.~\ref{eq:eq5}, it is calculated as the product of all component-wise intersections between emotion classes $i$ and $j$. The values for $I_{i,j}$ are in the range [0,1]. It has a maximum value of one when $i=j$, i.e. when the spread of $m^{th}$ component's data is the same for two emotion classes. Here, $I_{i,j}$ denotes the total intersection between the emotion classes $i$ and $j$. \vspace{-.3in} 

{\fontsize{10}{12}\selectfont 
\begin{eqnarray}
\label{eq:eq5}
\begin{split}	
I_{i,j}= I_{i,j}(x)*I_{i,j}(y)*I_{i,j}(z)
\end{split}
\end{eqnarray}
} \vspace{-.3in} 

\subsubsection{Comparison of Interpretability Approaches}
{
    The DnCShap technique compares with other interpretability techniques such as SHAP \cite{lundberg2017unified}, LIME \cite{ribeiro2016should}, and Grad-CAM \cite{selvaraju2017grad} as follows.\vspace{.1in}
}

\noindent {
    \textit{Computational Efficiency:} DnCShap enhances computational efficiency by incorporating a Divide \& Conquer strategy to compute Shapley values, reducing the time complexity from exponential, as is typical with traditional SHAP, to linear. Our method computes approximate Shapley values in linear time, compared to the quadratic or higher time complexity of SHAP and LIME.\vspace{.1in}
}

\noindent {
    \textit{Interpretability Depth:} Unlike Grad-CAM, which provides a coarse localization map indicative of relevant regions in the image, DnCShap offers detailed attributions at a pixel level, enabling finer interpretations of the contributions of specific image features to the emotion recognition task. This depth is particularly crucial in understanding subtle emotional cues in facial expressions.\vspace{.1in}
}

\noindent {
    \textit{Applicability to FER:} The ability of DnCShap to provide detailed and computationally efficient interpretations makes it particularly suited for real-time FER systems, where understanding the decision-making process of the model is as important as its accuracy. \vspace{-.2in}
}
\begin{table}[!h]
\centering
\caption{{Comparison of Interpretability Approaches.}}
\label{tab:icomparison}
\begin{tabular}{lccc}
\toprule
\textbf{Method} & \textbf{Computational Complexity} & \textbf{Detail of Interpretation} & \textbf{Applicability to FER} \\
\midrule
SHAP & High (Exponential) & High (Pixel-level) & Suitable \\
LIME & Variable (High) & Medium (Segment-level) & Suitable \\
Grad-CAM & Low (Constant) & Low (Region-level) & Limited \\
\textbf{DnCShap} & \textbf{Low (Linear)} & \textbf{High (Pixel-level)} & \textbf{Highly Suitable} \\
\bottomrule
\end{tabular}
\end{table}

{
    The observations, presented in Table \ref{tab:icomparison}, demonstrate that DnCShap not only outperforms existing methods in terms of computational efficiency but also provides deeper insights, making it an ideal choice for advanced FER applications where both speed and interpretability are crucial.\vspace{.1in}
}

\noindent {\textit{Computational Efficiency and Scalability}: DnCShap enhances computational efficiency by incorporating a Divide \& Conquer strategy, which significantly reduces the computational overhead associated with calculating Shapley values. Typically, computing exact Shapley values in traditional SHAP has exponential time complexity. In contrast, our method manages to approximate these values in linear time, which is particularly advantageous when scaling to more complex models or larger datasets. To demonstrate the scalability and runtime efficiency, we conducted benchmark tests across datasets of varying sizes and complexity, which show that DnCShap maintains efficient performance without compromising interpretability, even as dataset size increases. These tests help substantiate DnCShap's practical utility in real-world applications where both computational resources and time are limiting factors. }

\section{Experimental Results}\label{sec:implementation}
This section presents the implementation setup and experimental results along with ablation studies and a discussion of the important observations.

\subsection{Datasets}\label{sec:dataset1ch5b}
The datasets utilized in this study are characterized by their large scale and diversity, encompassing a broad range of emotions and scenarios. These qualities make them exceptionally well-suited for the demands of both facial and image emotion recognition.\vspace{.1in}

\noindent \textbf{FER Datasets}: The FER datasets used in this study are widely recognized in the research community for their robust and varied annotations of facial expressions, which are critical for the development and evaluation of FER systems. 
\begin{itemize}
    \item \textit{FER13}~\cite{carrier2013fer}: Includes 35,887 emotion-labeled images, utilized extensively for its diversity in facial expressions and occlusions, facilitating robust model training.
    \item \textit{FERG}~\cite{aneja2016modeling}: Comprises 55,767 annotated images known for their animated representations, which help in understanding the subtleties of facial expressions in animated contexts.
    \item \textit{JAFFE}~\cite{kamachi1998japanese}: Contains 213 images, often used for its high-quality depictions of basic emotional expressions by Japanese female models.
    \item \textit{CK+}~\cite{lucey2010extended}: Features 927 images with labeled emotions, widely used for its sequential imaging that captures the onset and apex of facial expressions.\vspace{.1in}
\end{itemize}

\noindent \textbf{IER Datasets}: These datasets are crucial for advancing research in image emotion recognition by providing a diverse range of contexts and settings.
\begin{itemize}
    \item \textit{IAPSa} dataset \cite{mikels2005emotional}: Comprising 395 images, this dataset is pivotal for its role in psychophysiological studies and is frequently used to validate emotion recognition models due to its reliable emotional annotations.
    \item \textit{ArtPhoto} dataset \cite{machajdik2010affective}: Contains 806 art photos labeled with emotions by the artists, offering a unique dataset that bridges art and emotion, providing insights into the aesthetic elements that evoke emotional responses.
    \item \textit{Flicker \& Instagram (FI)} dataset \cite{you2016building}: With 23,308 images, it represents the largest dataset used in our study, showcasing a wide variety of real-world scenarios that are instrumental in developing robust IER systems.
    \item \textit{EMOTIC} dataset \cite{emotic_pami2019}: Features a comprehensive collection of 23,571 images in diverse contexts, labeled with both continuous dimensions and discrete categories, making it invaluable for understanding emotional expressions in naturalistic settings.
\end{itemize}

\subsection{Emotion Class Maping and Hyperparameter Setting}
\subsubsection{Emotion Class Mapping} \label{sec:class_mapping}
{
In this work, we consolidate all emotion annotations to four discrete classes, i.e., \emph{happy}, \emph{sad}, \emph{hate}, and \emph{anger}. We chose these four because they are the most frequently used categories in state-of-the-art Image Emotion Recognition (IER) studies involving the same benchmark datasets: IAPSa, ArtPhoto, Flicker\&Instagram (FI), and EMOTIC. As discussed in Section~\ref{sec:compare}, many prior methods on these datasets also focus on these four dominant categories for consistency and reliable sample sizes.\vspace{.1in}}

\noindent {\textit{Rationale for Merging}:
Although some datasets originally contain up to seven labeled emotions (e.g., \emph{amusement}, \emph{contentment}, \emph{disgust}, \emph{fear}, \emph{surprise}, etc.), not all of these are well-represented or consistently annotated across the four datasets. For instance, IAPSa and ArtPhoto include additional labels like \emph{amusement} or \emph{contentment}, but each has relatively few samples. Likewise, \emph{disgust} is labeled in some, but not all. To avoid excessive class imbalance and enhance cross-dataset alignment, we adopt the following merges aligned with Plutchik’s emotion wheel~\cite{plutchik2001nature}: \emph{contentment} and \emph{amusement} are merged to \emph{happy} whereas \emph{disgust} is merged to \emph{hate}. In this manner, each of the four datasets has sufficient examples in the four selected emotion categories, facilitating robust training and evaluation.\vspace{.1in}}

\noindent {\textit{Impact on Model Performance}:
By consolidating underrepresented labels into these four commonly used categories, we mitigate the risk of overfitting to sparsely populated classes and reduce confusion between closely related affective states (e.g., \emph{disgust} vs.\ \emph{hate}). Empirically, we observe more stable training curves and higher overall accuracy compared to attempts at using six or seven fine-grained categories. In particular, merging \emph{disgust} into \emph{hate} and grouping \emph{contentment/amusement} with \emph{happy} provides clearer class boundaries, resulting in better cross-dataset generalization. While we focus on these four discrete classes for the present work, the underlying domain adaptation framework could be extended to additional discrete labels if future datasets offer consistent and balanced annotations.}

\subsubsection{Training Strategy and Hyperparameter Setting}
The model training is implemented in PyTorch, utilizing a training-testing data split of $80\%$ and $20\%$, $5$-fold cross-validation, with a standard batch size of 16, and a learning rate ranging from $8 \times 10^{-5}$ to $8 \times 10^{-4}$ for up to $100$ epochs. The FER and IER models were initially trained using cross-entropy loss and subsequently together using the discrepancy loss, with the Adam optimizer utilized for both networks. Accuracy, defined as the proportion of true results (both true positives and true negatives) among the total number of cases examined, has been used as the evaluation metric.\vspace{.1in}

\noindent {\textit{Hyperparameter Tuning}: We employed Ray Tune \cite{liaw2018tune} for hyperparameter optimization, allowing for a systematic and efficient search through the hyperparameter space. This approach facilitated the adjustment of learning rates dynamically, refinement of batch sizes to balance computational efficiency and model accuracy, and tuning of Adam optimizer parameters to optimize training dynamics. Different hyperparameter settings, such as beta values and epsilon for the Adam optimizer, were methodically varied to explore their impacts on the loss oscillations and the stability of model updates. Extended training durations beyond the standard $100$ epochs with early stopping based on validation loss were also tested to prevent overfitting while ensuring robust model performance. Inspired by other advanced hyperparameter optimization tools such as Optuna \cite{akiba2019optuna} and Keras Tuner \cite{omalley2019kerastuner}, which are used extensively in various machine learning platforms, we incorporated similar strategies within our Ray Tune implementation to fine-tune the balance between cross-entropy and discrepancy loss. These modifications aided in refining the model’s capability to learn generalized features while effectively minimizing domain-specific discrepancies.}
 
\subsection{Models}\label{sec:models} 
The following baselines and proposed system's architectures have been decided according to the ablation studies discussed in Section \ref{sec:ablationch5b}.

\begin{itemize} 	\vspace{.07in}
\item \textbf{Baseline\#1}: It trains AlexNet \cite{krizhevsky2017imagenet} simultaneously for FER and IER.\vspace{.07in}  

\item \textbf{Baseline\#2}: The VGG16~\cite{simonyan2014very} based IER and FER models are implemented and trained using discrepancy loss. \vspace{.07in}

\item \textbf{Baseline\#3}: It trains ResNet~\cite{he2016deep} models simultaneously for FER and IER. \vspace{.07in}

\item \textbf{Baseline\#4}: It generates the captions for the given images and trains the TER model for emotion recognition. The captions' emotion labels are considered the images' labels, as there is a one-to-one mapping between them. \vspace{.07in}

\item \textbf{Baseline\#5}: It adapts a pre-trained FER model, Deep Emotion \cite{minaee2021deep} and re-trains it simultaneously for FER and IER using the discrepancy loss. \vspace{.07in}
{
    \item \textbf{Baseline\#6}: uses EfficientNet \cite{tan2019efficientnet}. It is a scalable architecture that aims to achieve superior efficiency through compound scaling of depth, width, and resolution. \vspace{.07in}
}{
    \item \textbf{Baseline\#7}: It is based on the Vision Transformer (ViT) \cite{dosovitskiy2020image} model that applies attention mechanisms to segment the images into patches and processes them sequentially. \vspace{.07in}
}{
    \item \textbf{Baseline\#8} uses Swin Transformer \cite{liu2021swin} which is a hierarchical transformer whose representation is computed with shifted windows, facilitating efficient modeling of image data at multiple scales to handle complex visual tasks. \vspace{.07in}
}

\item \textbf{Proposed system}: The proposed IER system adapts the FER model proposed in section \ref{sec5.1.2} using the proposed domain adaptation scheme. 
\end{itemize} 

\subsection{Results}\label{sec:resultsch5b}
This section discusses the IER results along with the interpretable feature maps. 

\subsubsection{Accuracy and Confusion Matrices}
The proposed IER system shows 61.86\%, 62.47\%, 70.78\%, and 59.72\% accuracies for IAPSa, ArtPhoto, FI, and EMOTIC datasets, respectively. Fig. \ref{fig:cm} shows the corresponding confusion matrices. 

\begin{figure}[]
\centering
\includegraphics[width=.9\textwidth]{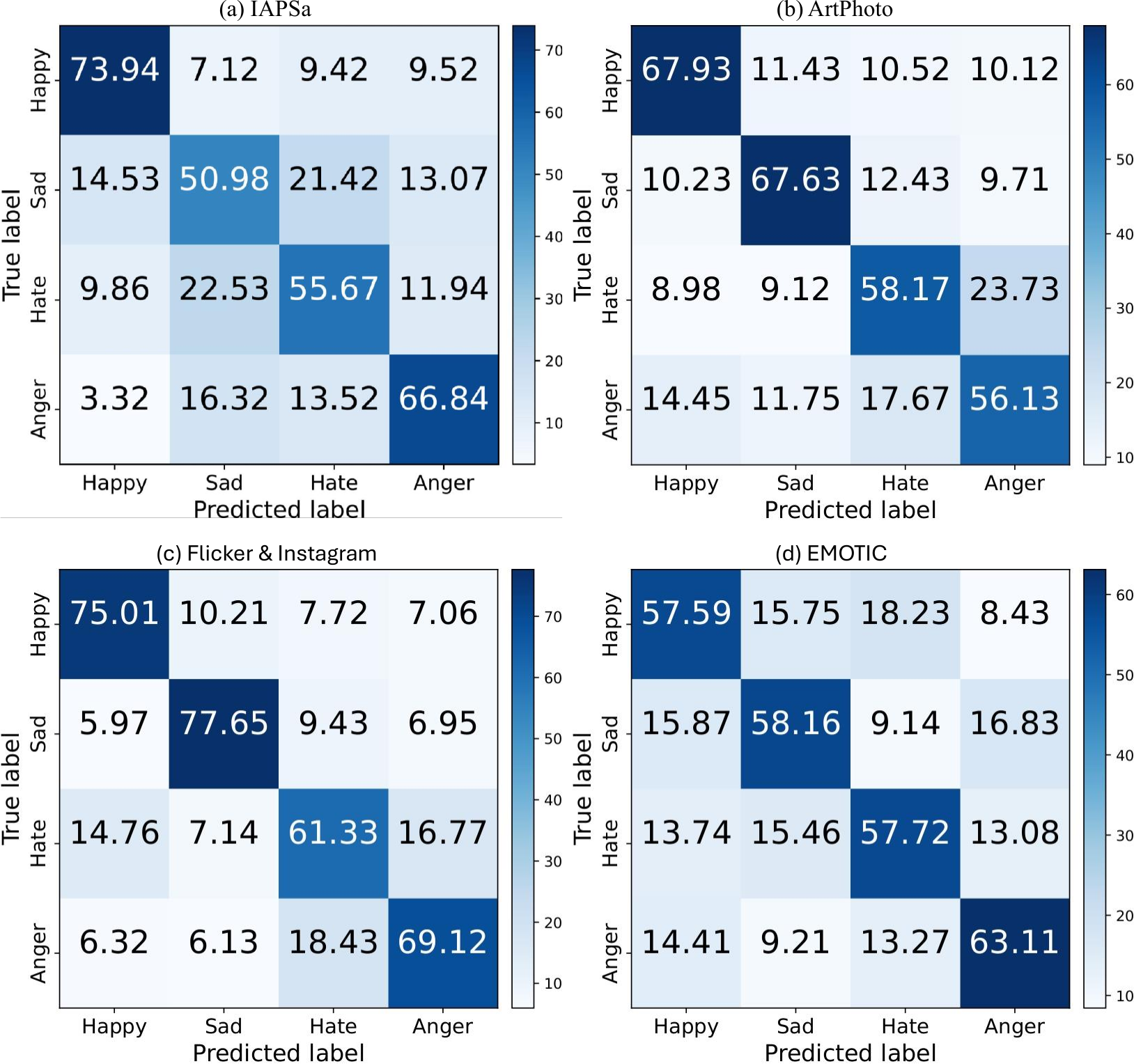} 
\caption{Confusion matrices for IAPSa, ArtPhoto, FI, and EMOTIC datasets.}
\label{fig:cm}
\end{figure}

\subsubsection{Comparison With Existing Methods and Baselines}\label{sec:compare}
Table~\ref{tab:ier_sota} compares the results of the proposed IER system with the existing methods and baseline models. Several research works utilize diverse evaluation metrics including Mean Average Precision (mAP), Area Under the Curve (AUC), and accuracy percentages. For consistency, this analysis focuses on studies that use accuracy as their evaluation metric, ensuring a uniform standard for comparing results across various IER methods and datasets. These datasets predominantly feature four common emotion classes (`happy,' `anger,' `hate,' and `sad'), aligning with the typical categorizations found in existing IER research.

\begin{table}[]
\centering
\caption{{Comparison of proposed system's results with baselines and existing methods, across various datasets. The best and second-best results are highlighted in bold and underline, respectively. Standard deviations from five-fold cross-validation are shown in parentheses to indicate error margins.}}
\label{tab:ier_sota}
\resizebox{1\textwidth}{!}
{
\begin{tabular}{lcccc}
\toprule
\textbf{Model} & \textbf{IAPSa} & \textbf{ArtPhoto} & \textbf{FI} & \textbf{EMOTIC}\\ 
\midrule
Baseline\#1 & 42.83\% (±1.5\%) & 47.37\% (±1.2\%) & 58.21\% (±1.8\%) & 41.17\% (±2.0\%) \\ 
Baseline\#2 & 48.87\% (±1.3\%) & 50.63\% (±1.4\%) & 59.74\% (±1.6\%) & 45.45\% (±1.9\%) \\ 
Baseline\#3 & 48.31\% (±1.6\%) & 52.26\% (±1.1\%) & 63.17\% (±1.3\%) & 46.57\% (±1.7\%) \\ 
Baseline\#4 & 53.72\% (±1.2\%) & 50.42\% (±1.5\%) & 59.45\% (±1.4\%) & 49.32\% (±1.6\%) \\ 
Baseline\#5 & 57.46\% (±1.0\%) & 55.87\% (±1.3\%) & 63.12\% (±1.1\%) & 54.37\% (±2.1\%) \\ 
{Baseline\#6} & 57.81\% (±0.9\%) & 54.62\% (±1.2\%) & 66.72\% (±1.0\%) & \underline{58.25\% (±2.0\%)} \\ 
{Baseline\#7} & 59.13\% (±1.1\%) & 55.24\% (±1.4\%) & 65.86\% (±1.2\%) & 57.65\% (±1.8\%) \\ 
{Baseline\#8} & 58.87\% (±1.0\%) & 53.18\% (±1.3\%) & 67.12\% (±1.3\%) & 57.54\% (±1.5\%)\vspace{.03in}\\ \hdashline 
Deep Metric Learning \cite{yang2018retrieving} & 44.20\% & 40.00\% & - & - \\
Art Theory Features \cite{machajdik2010affective} & - & 49.50\% & - & - \\
Feature-based IER \cite{zhao2014exploring} & 49.46\% & 51.72\% & - & - \\ 
Instance Learning \cite{rao2016multi} & - & - & 51.67\% & - \\
{Fuzzy-aware Loss} \cite{zheng2025fuzzy} & - & - & 57.20\% & - \\
Fine-tuned CNN \cite{you2016building} & - & - & 58.30\% & - \\
MldrNet \cite{rao2019learning} & - & - & 67.75\% & - \\
Context-based Analysis \cite{emotic_pami2019} & - & - & - & 42.02\% \\
One-shot Learning \cite{peng2021affect} & - & - & - & 54.89\% \\
Multi-instance Learning \cite{rao2019learning} & \underline{59.93\%} & - & - & - \\
{Multi-task DA} \cite{zhao2016predicting} & - & 60.36\% & 63.92\% & - \\
{CycleGAN} \cite{zhu2017unpaired} & - & 61.11\% & 63.87\% & - \\
{CycleEmotionGAN} \cite{zhao2019cycleemotiongan} & - & 61.96\% & 67.78\% & - \\
{Deep Cocktail Network} \cite{xu2018deep} & - & 62.34\% & 65.31\% & - \\
{Multi-source DA} \cite{lin2020multi} & - & \textbf{63.58\%} & \underline{70.63\%} & - 
\vspace{.03in}\\
\hdashline
\textit{Proposed} & \textbf{61.86\% (±0.8\%)} & \underline{62.47\% (±1.0\%)} & \textbf{70.78\% (±1.1\%)} & \textbf{59.72\% (±1.2\%)} \\
\bottomrule
\end{tabular}
}
\end{table}

\subsubsection{Feature-wise Interpretability}
Fig. \ref{fig:fmapsIER} demonstrates the feature maps for various emotion classes. The maps highlighted in red illustrate the most significant visual features that contribute to emotion recognition. These maps are instrumental in identifying both facial and non-facial cues that are pivotal for accurately classifying emotions. For images containing human faces, the feature maps effectively capture distinct facial features such as eyebrows, eyes, and mouth, which are crucial for recognizing emotions like happiness or anger. In contrast, for images without human faces, the feature maps help in pinpointing other relevant areas, such as body posture or contextual elements within the scene, that are significant in conveying the underlying emotion of the image. This detailed visualization aids in understanding how different components of an image contribute to emotion recognition, providing valuable insights into the robustness and adaptability of the IER system in handling diverse types of images.

\begin{figure}[]
\centering
\subfigure[`Happy']
{
\begin{tabular}{cccc}
{\resizebox{2.1cm}{!}{\includegraphics[width=\textwidth]{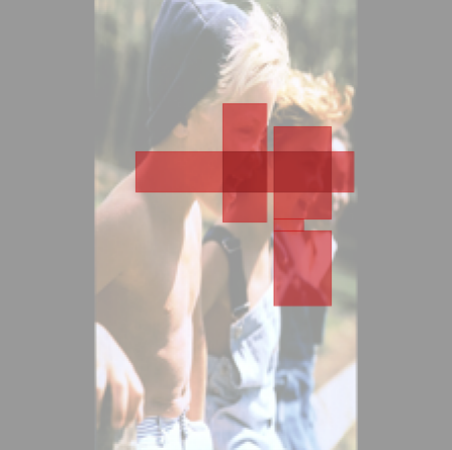}}}
&{\resizebox{2.1cm}{!}{\includegraphics[width=\textwidth]{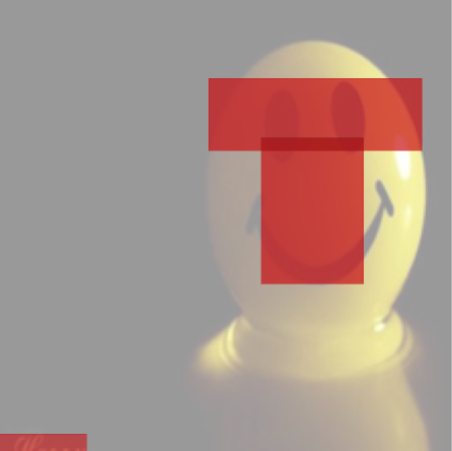}}}
&{\resizebox{2.1cm}{!}{\includegraphics[width=\textwidth]{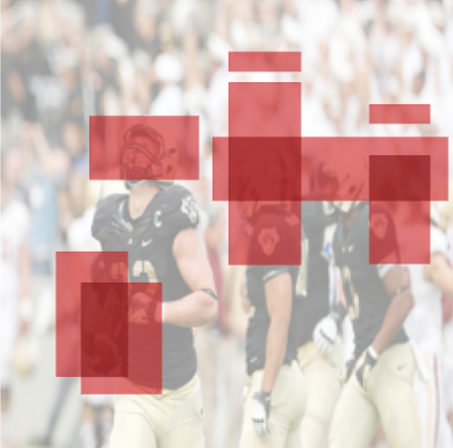}}}
&{\resizebox{2.1cm}{!}{\includegraphics[width=\textwidth]{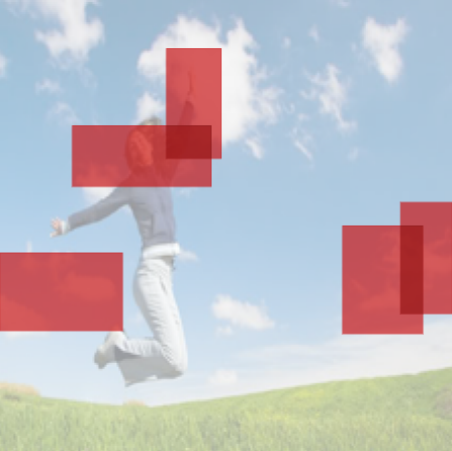}}}\\
IAPSa &ArtPhoto &FI &EMOTIC
\end{tabular} 
}\\ 
\subfigure[`Sad']
{
\begin{tabular}{cccc}
{\resizebox{2.1cm}{!}{\includegraphics[width=\textwidth]{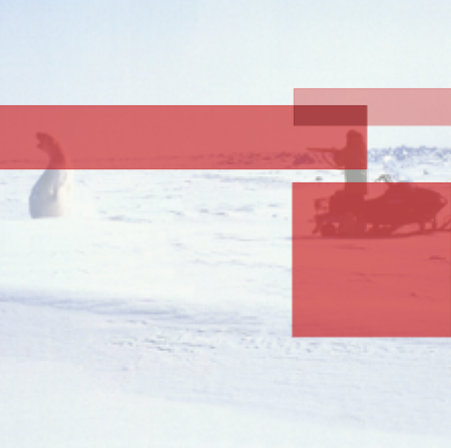}}}
&{\resizebox{2.1cm}{!}{\includegraphics[width=\textwidth]{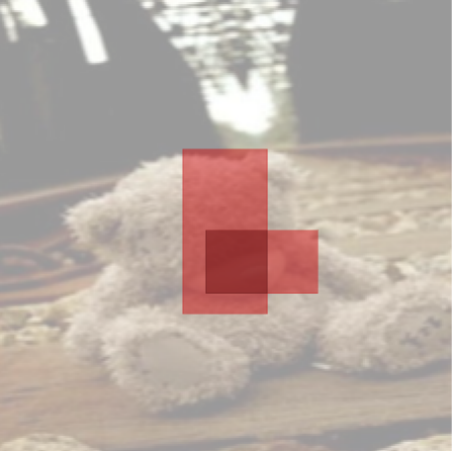}}}
&{\resizebox{2.1cm}{!}{\includegraphics[width=\textwidth]{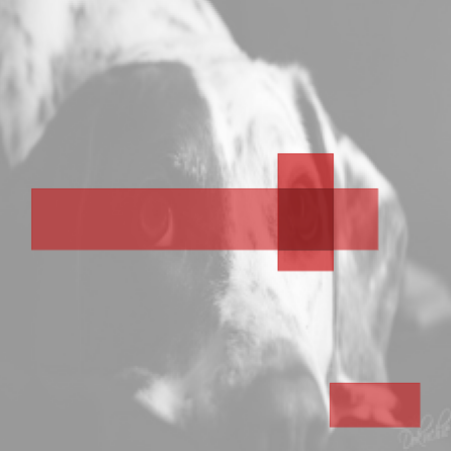}}}
&{\resizebox{2.1cm}{!}{\includegraphics[width=\textwidth]{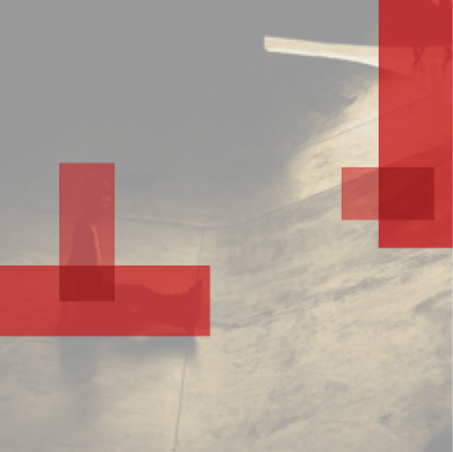}}}\\
IAPSa &ArtPhoto &FI &EMOTIC
\end{tabular} 
}\\ 
\subfigure[`Hate']
{
\begin{tabular}{cccc}
{\resizebox{2.1cm}{!}{\includegraphics[width=\textwidth]{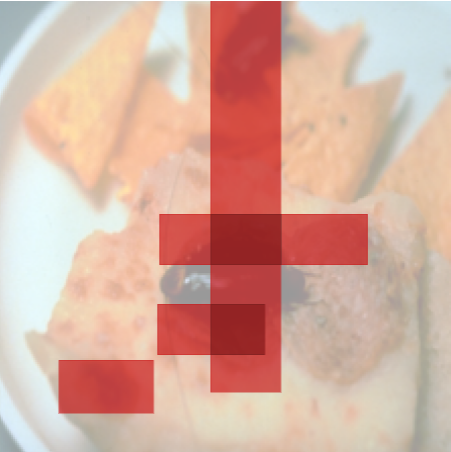}}}
&{\resizebox{2.1cm}{!}{\includegraphics[width=\textwidth]{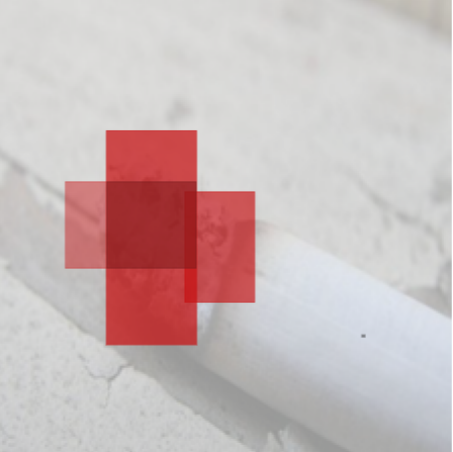}}}
&{\resizebox{2.1cm}{!}{\includegraphics[width=\textwidth]{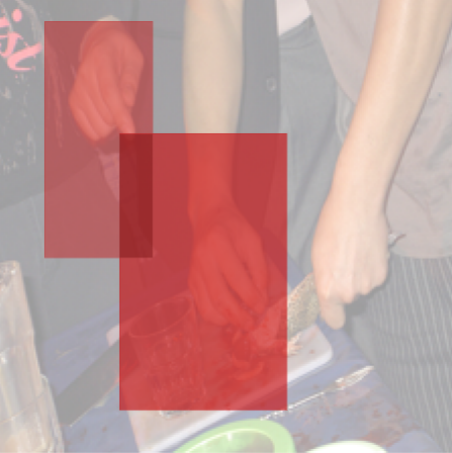}}}
&{\resizebox{2.1cm}{!}{\includegraphics[width=\textwidth]{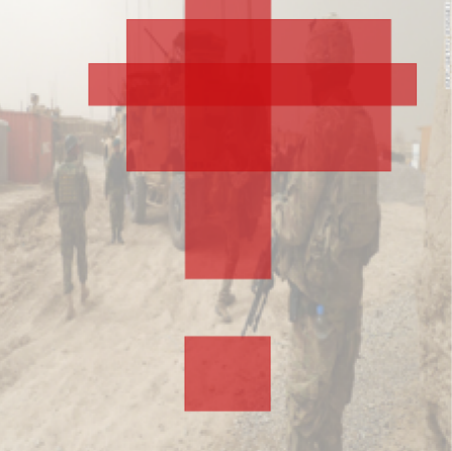}}}\\
IAPSa &ArtPhoto &FI &EMOTIC
\end{tabular} 
}\\ 
\subfigure[`Anger']
{ 
\begin{tabular}{cccc}
{\resizebox{2.1cm}{!}{\includegraphics[width=\textwidth]{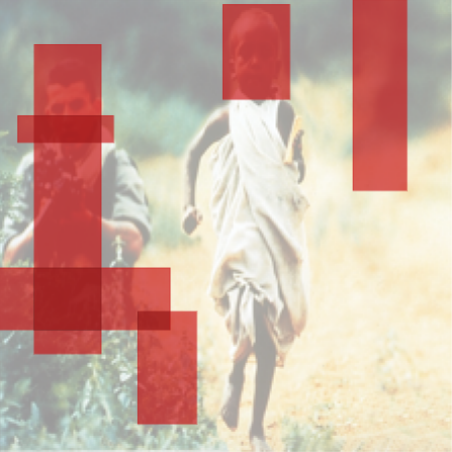}}}
&{\resizebox{2.1cm}{!}{\includegraphics[width=\textwidth]{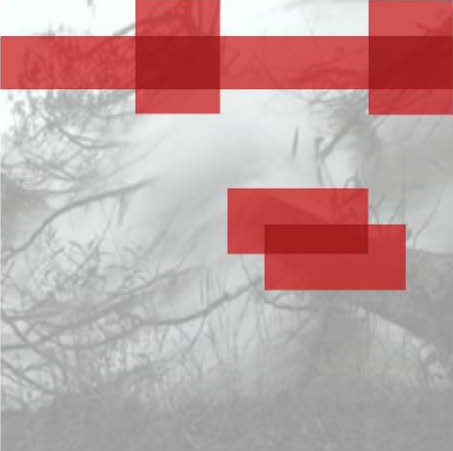}}}
&{\resizebox{2.1cm}{!}{\includegraphics[width=\textwidth]{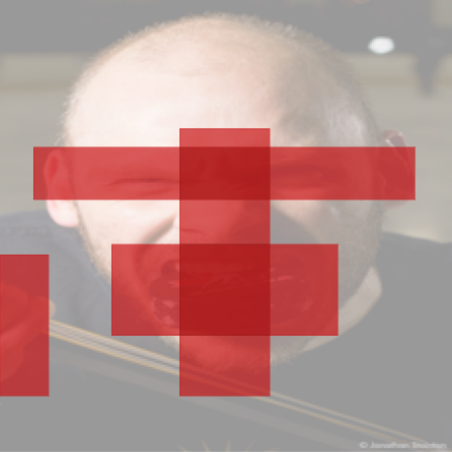}}}
&{\resizebox{2.1cm}{!}{\includegraphics[width=\textwidth]{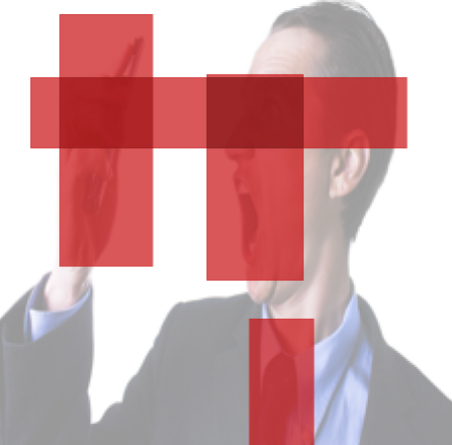}}}\\
IAPSa &ArtPhoto &FI &EMOTIC
\end{tabular} 
}~\\
\caption{Feature-wise Interpretability for various emotion classes. Areas marked in red highlight the most significant visual features contributing to emotion recognition.}  
\label{fig:fmapsIER}
\end{figure}

\subsubsection{Layer-wise Interpretability}
Fig. \ref{fig:emb} illustrates the embeddings from the trained IER model, projected onto a three-dimensional hyperplane. The principal components of the embeddings are spread along the x, y, and z axes, representing distinct dimensions of the data's variance for various emotion classes. This visualization highlights how different layers of the model capture varying levels of abstraction in the data. Notably, as we progress from the intermediate to the final layers of the network, the separation between the embeddings of different emotion classes becomes more pronounced. This enhanced separation indicates that the deeper layers of the network are more effective at distinguishing between complex emotional states, suggesting that these layers capture more discriminative features essential for accurate emotion classification. This layer-wise analysis provides valuable insights into the model’s ability to generalize and its interpretability, demonstrating how different network depths contribute to the overall decision-making process.

\begin{figure}[]
\centering\vspace{-.08in}
\subfigure[IAPSa]
{
\begin{tabular}{ccc}
{\resizebox{3.75cm}{!}{\includegraphics[width=\textwidth]{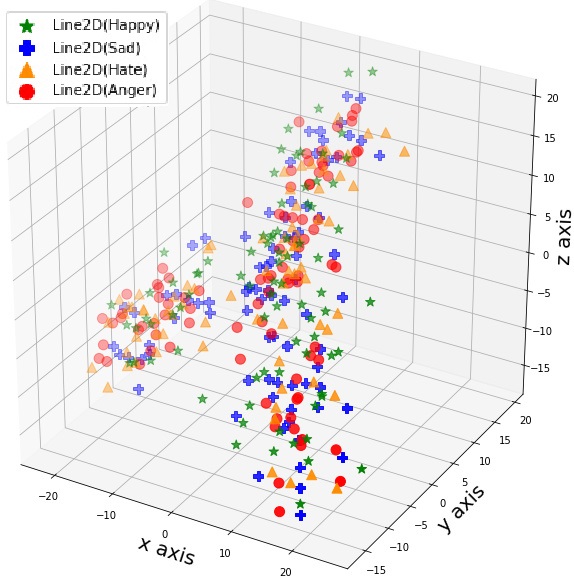}}}
&{\resizebox{3.75cm}{!}{\includegraphics[width=\textwidth]{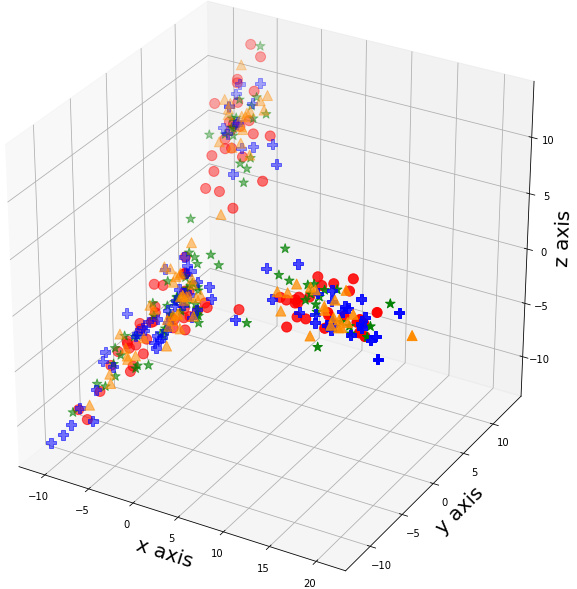}}}
&{\resizebox{3.75cm}{!}{\includegraphics[width=\textwidth]{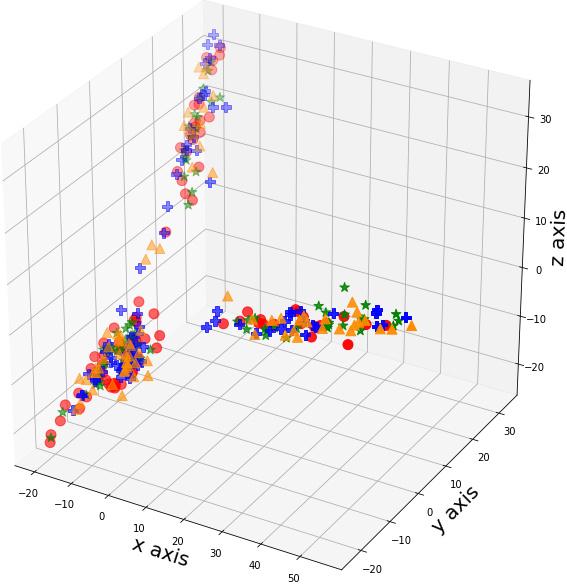}}}\\
Third Last Layer &Second Last Layer &Output Layer 
\end{tabular}  
}~\\\vspace{-.05in}
\subfigure[ArtPhoto]
{
\begin{tabular}{ccc}
{\resizebox{3.75cm}{!}{\includegraphics[width=\textwidth]{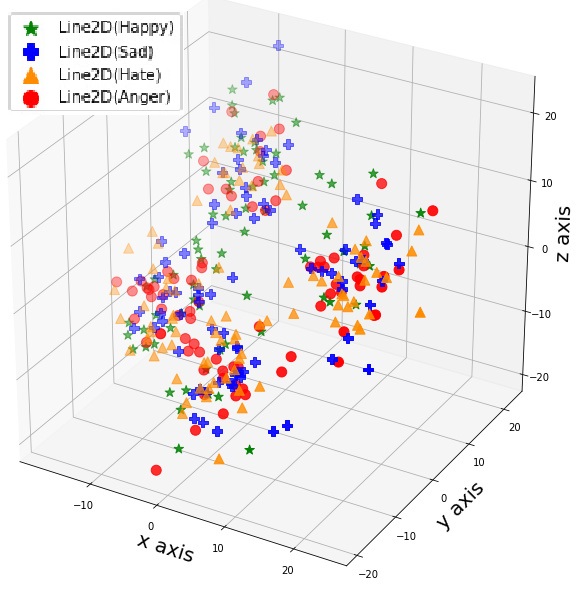}}}
&{\resizebox{3.75cm}{!}{\includegraphics[width=\textwidth]{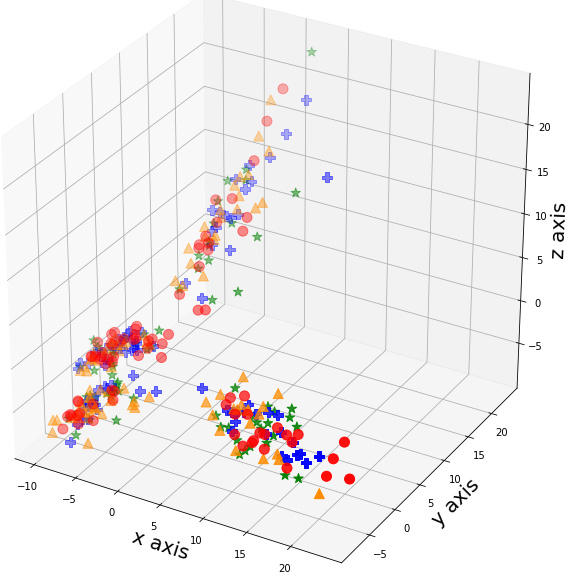}}}
&{\resizebox{3.75cm}{!}{\includegraphics[width=\textwidth]{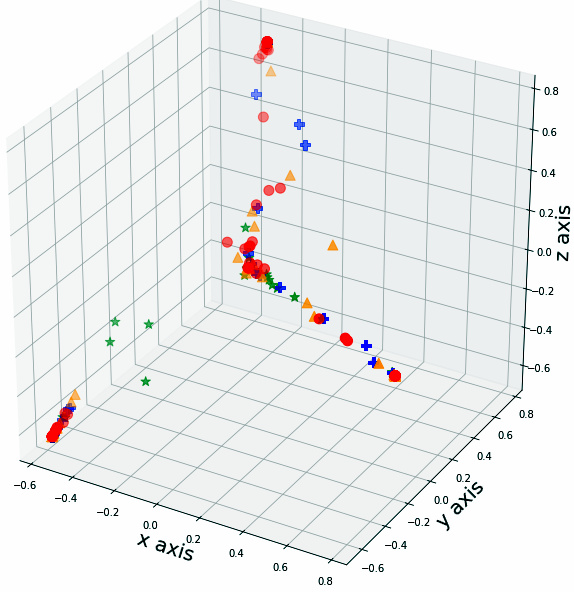}}}\\
Third Last Layer &Second Last Layer &Output Layer 
\end{tabular}  
}~\\\vspace{-.05in}
\subfigure[FI]
{
\begin{tabular}{ccc}
{\resizebox{3.75cm}{!}{\includegraphics[width=\textwidth]{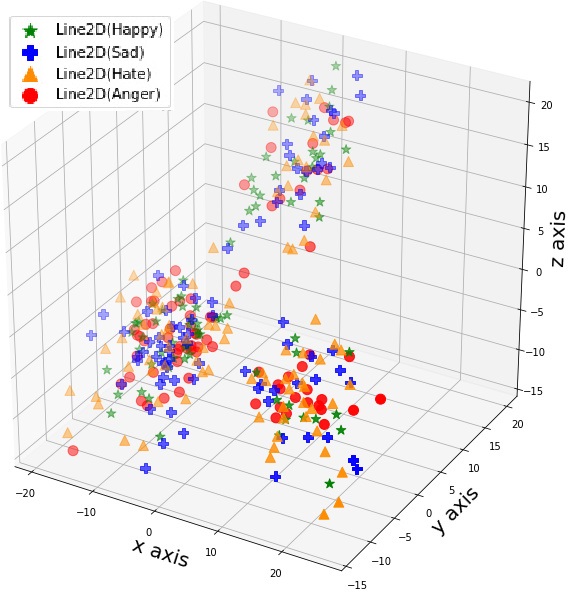}}}
&{\resizebox{3.75cm}{!}{\includegraphics[width=\textwidth]{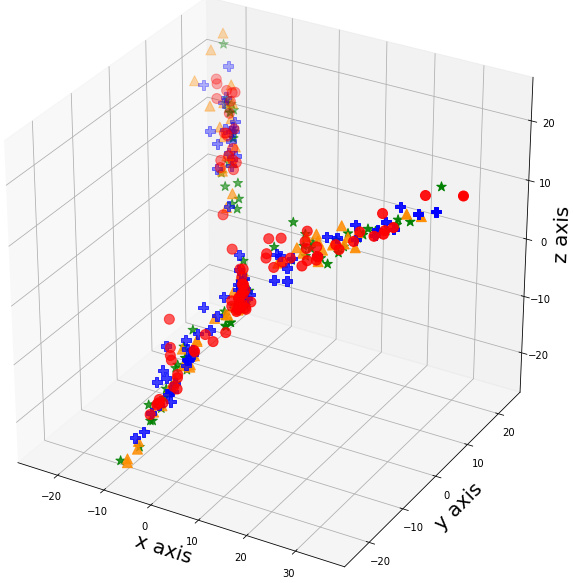}}}
&{\resizebox{3.75cm}{!}{\includegraphics[width=\textwidth]{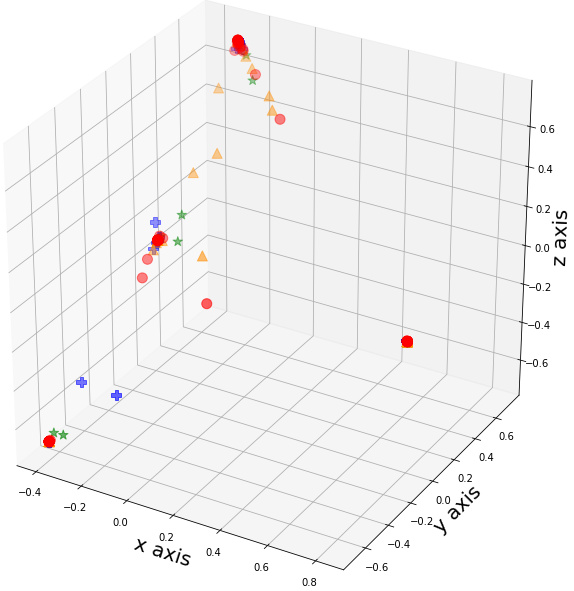}}}\\
Third Last Layer &Second Last Layer &Output Layer 
\end{tabular}  
}~\\\vspace{-.05in}
\subfigure[EMOTIC]
{
\begin{tabular}{ccc}
{\resizebox{3.75cm}{!}{\includegraphics[width=\textwidth]{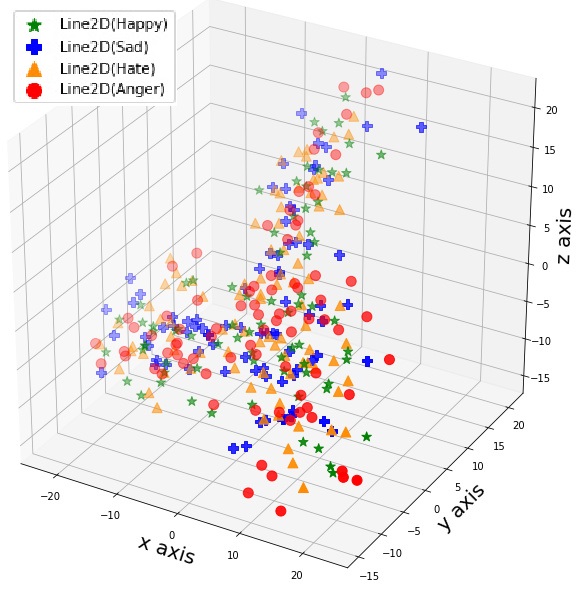}}}
&{\resizebox{3.75cm}{!}{\includegraphics[width=\textwidth]{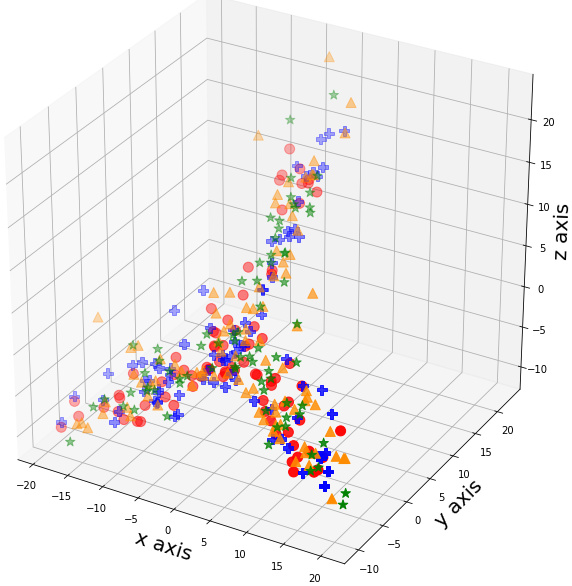}}}
&{\resizebox{3.75cm}{!}{\includegraphics[width=\textwidth]{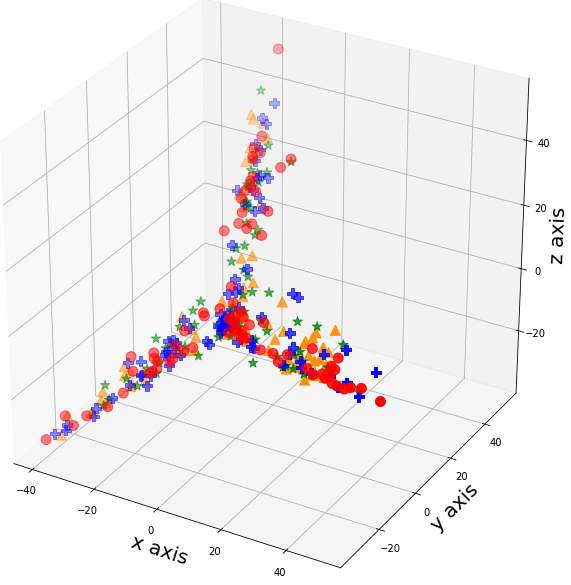}}}\\
Third Last Layer &Second Last Layer &Output Layer 
\end{tabular}  
} 
\caption{Layer-wise Interpretability for various IER datasets, projecting the embeddings from the trained IER model onto a three-dimensional hyperplane across various layers. This visualization illustrates how each layer contributes to the differentiation and recognition of emotional states, providing insights into the depth-specific processing capabilities of the model.}  
\label{fig:emb}
\end{figure} 

\subsubsection{Computational Efficiency Analysis}\label{sec:comp_eff}
{
    Table~\ref{tab:ablationch6} shows per-epoch computational times on an RTX~2080 GPU with a batch size of 16. We analyze four main operations in our pipeline: \textit{Baseline Training:} The time to train the baseline model on each dataset for one epoch without domain adaptation. \textit{Domain Adaptation:} The overhead introduced by our discrepancy-loss--based alignment between FER and IER, added at the end of each epoch. \textit{DnCShap:} The additional cost for generating the Shapley-based attributions with Divide \& Conquer Shap (DnCShap) each epoch. \textit{PCA Embeddings:} Time to extract embeddings from the final layer and perform PCA for interpretability analyses.
}

\begin{table}[!h] 
\centering 
{\fontsize{10}{12}\selectfont 
\caption{{Per-epoch Computational Cost Analysis. Baseline training refers to the cost of plain supervised learning, while Domain Adaptation shows the extra overhead of our discrepancy-based alignment. Feature and Layer-wise Interpret reflect the per-epoch time to produce interpretability outputs. Increase percentages compared to the baseline are shown with an upward arrow.}} 
\label{tab:ablationch6}
\resizebox{1\textwidth}{!}
{
\begin{tabular}{cccccc}
\toprule
\textbf{Component} & \textbf{Operation} & \textbf{IAPSa} & \textbf{ArtPhoto} & \textbf{FI} & \textbf{EMOTIC} \\ 
\midrule
\textbf{Baseline Training} & Training Time (s) & 4 & 6 & 25 & 28 \\
\textbf{Domain Adaptation} & Adaptation Time (s) & 1.2 & 1.9 & 7.2 & 9.3 \\
\textbf{Feature-wise Interpret} & Analysis Time (s) & 0.5 & 0.6 & 2.8 & 3.2 \\
\textbf{Layer-wise Interpret} & Analysis Time (s) & 0.3 & 0.4 & 1.7 & 1.9 \\
\hdashline
\textbf{Total} & Overall (s) & 6.0 (50\%$\uparrow$) & 8.9 (48\%$\uparrow$) & 36.7 (47\%$\uparrow$) & 42.4 (51\%$\uparrow$) \\
\bottomrule
\end{tabular}
}}\vspace{.1in}
\end{table}

{
    We observe that smaller datasets (IAPSa and ArtPhoto) require only a few seconds per epoch, whereas larger datasets (FI and EMOTIC, each containing over 20,000 images) demand correspondingly more time. Despite the discrepancy-loss domain adaptation overhead, it remains only a modest fraction (\(\sim\)20--40\%) of baseline training. DnCShap analysis takes a few seconds at most, owing to its linear-time complexity. PCA computations are also minimal. Overall, even with additional domain adaptation and interpretability steps, our method remains computationally tractable for practical use, completing training in a reasonable time (on the order of minutes per epoch) for the smaller datasets and under a minute per epoch for the larger FI and EMOTIC datasets.
}

\subsubsection{Human Annotation Study}
\label{sec:human-study}
{To evaluate how well our model’s interpretations align with human perception of non-facial or generic images, we conducted a small-scale annotation study on a subset of each dataset (IAPSa, ArtPhoto, FI, and EMOTIC). We randomly selected 100 images per dataset that primarily depict scenes, objects, or contexts without clear human faces.\vspace{.1in}}

\noindent {\textit{Annotation Procedure}:
Three volunteers (postgraduate students with familiarity in affective computing) independently labeled each image with one of the four emotions (\emph{happy, sad, hate, anger}) and indicated the visual cues driving their label choice. They received no information about model outputs.\vspace{.1in}}

\noindent {\textit{Inter-Annotator Agreement}:
We used Cohen’s Kappa ($\kappa$) \cite{vieira2010cohen} to quantify pairwise agreement. Table~\ref{tab:human_study} shows the average $\kappa$ across annotator pairs. The moderate agreement for FI and EMOTIC underscores the subjective nature of emotional cues in real-world or context-heavy scenes.\vspace{.1in}}

\noindent {\textit{Comparison with Model}:
We compared the model’s predicted labels to the majority vote among annotators for each image. In 78--83\% of cases, the model matched the humans’ dominant label. Additionally, DnCShap-highlighted regions overlapped with human-indicated cues (e.g., color or foreground objects) in 70\% of instances.}

\begin{table}[!h]
\centering
\caption{{Human Annotation Study on 50 Non-Facial Images per Dataset. \emph{Model-Human Align} is the proportion of images where the model prediction matched the majority human label.}}
\label{tab:human_study}
\begin{tabular}{lccc}
\toprule
\textbf{Dataset} & \textbf{Avg. Cohen’s ($\kappa$)} & \textbf{Model-Human Align} & \textbf{Annotators’ Agreement}\\
\midrule
IAPSa     & 0.79 & 81.16\% & Moderate-High\\
ArtPhoto  & 0.76 & 78.65\% & Moderate\\
FI        & 0.81 & 83.23\% & High\\
EMOTIC    & 0.75 & 79.64\% & Moderate\\
\bottomrule
\end{tabular}
\end{table}

{Overall, this study indicates that our model’s predictions and saliency outputs provide a reasonable correspondence with human judgments in generic, non-facial images. However, the moderate agreement levels among annotators also highlight the inherent subjectivity in emotional perception. Future extensions could expand the subset size and include a broader pool of annotators or alternative emotion categories.
}

\subsubsection {Ablation Studies}\label{sec:ablationch5b}
This section discusses the experimental studies performed to determine the appropriate domain adaptation approach.\vspace{.1in}

\noindent \textit{a) Inter-Modality Domain Adaptation: }
This trial transforms the visual information into textual and performs the emotion recognition task. This transformation has been performed using image captions generated using an attention-based captioning system~\cite{xu2015show}. Following this, a BERT based pre-trained Text Emotion Recognition (TER) model has been re-trained using these captions, and TER has been performed. The predicted TER labels have been considered the IER labels because the images and captions have one-to-one relations as they verbalize the same information visually and textually. The adapted TER model showed 33.89\% accuracy on being tested without re-training on image captions. On the other hand, re-training it with image captions demonstrated 53.07\% accuracy. The above-described approach has been implemented in Baseline\#4.\\

\noindent \textit{b) Intra-Modality Adaptation: }The datasets and pre-trained models for visual object recognition and FER tasks are more abundant. The following approaches have been considered to leverage them for IER. 

\begin{itemize}
\item 
The pre-trained object recognition models such as AlexNet, ResNet, and VGG16 were used for IER without re-training. However, it resulted in poor performance of around 30\% emotion recognition accuracy.  

\item
The pre-trained FER models were trained further and used for recognizing emotions in images using transfer learning. Taking inspiration from the existing feature transformation-based domain adaptation approaches \cite{saenko2010adapting}, we have mixed a fraction of IER data with FER data while training the FER model proposed in section \ref{sec5.1.2}. It enables the FER model to learn the distribution of IER data gradually. Table \ref{tab:ablationch5b} shows the IER performance on mixing a fraction of IER data along with FER data to train the FER models and test them for IER data. It has been observed that mixing 40-80\% IER data while training the FER model resulted in the best possible IER accuracies by this approach. However, there is a scope to improve it further. 

\begin{table}[]
\centering
{\fontsize{10}{12}\selectfont
\caption[Transformation-based IER experiments using FER and IER datasets.]{Transformation-based IER experiments, where a fraction \(x_i\) of the IER dataset is mixed with the FER dataset to train the FER model and test it on the IER dataset. The best and second-best results are highlighted in bold and underline, respectively. This table illustrates how varying proportions of IER data integrated into FER training affect the performance on IER tasks.} 
\label{tab:ablationch5b}
\resizebox{1\textwidth}{!}
{%
\begin{tabular}{lcccccccccc}\toprule 	
    \multirow{2}{*}{\pbox{10cm}{\textbf{FER \newline dataset}}} & \multicolumn{10}{c}{\textbf{$\mathbf{x_1}$ value for IAPSa IER dataset}} \\ \cmidrule{2-11} 
    & 0.1 & 0.2 & 0.3 & 0.4 & 0.5 & 0.6 & 0.7 & 0.8 & 0.9 & 1.0 \\ \midrule
    \textbf{FER13}   &36.68\% &39.33\% &43.25\% &46.65\% &47.61\% &48.61\% &48.16\% &49.84\% &\underline{50.53\%} &48.95\%\\
    \textbf{FERG}    &22.53\% &26.55\% &33.47\% &37.02\% &40.11\% &44.32\% &43.19\% &43.22\% &42.32\% &40.67\%\\
    \textbf{JAFFE}   &28.46\% &32.35\% &37.40\% &41.81\% &41.16\% &40.53\% &39.53\% &38.28\% &38.62\% &39.14\%\\
    \textbf{CK+}     &34.19\% &39.41\% &42.78\% &43.58\% &43.30\% &47.74\% &49.69\% & \textbf{51.21}\% &48.11\% &49.99\%\\ \bottomrule	
    & & & & & & & & & &\\ \toprule 	
    
    \multirow{2}{*}{\pbox{10cm}{\textbf{FER \newline dataset}}}&\multicolumn{10}{c}{\textbf{$\mathbf{x_2}$ value for ArtPhoto IERdataset}}\\\cmidrule{2-11}
    &0.1&0.2&0.3&0.4&0.5&0.6&0.7&0.8&0.9&1.0\\\midrule
    \textbf{FER13}	&31.74\% &38.31\% &39.99\% &42.61\% &45.69\% &45.21\% &45.13\% &45.99\% &45.44\% &44.98\%\\
    \textbf{FERG}	&29.91\% &36.44\% &40.09\% &41.13\% &43.82\% &42.01\% &43.49\% &42.15\% &42.41\% &42.86\%\\
    \textbf{JAFFE}	&37.26\% &39.61\% &41.96\% &45.94\% &46.62\% &45.61\% &45.61\% &46.16\% &\textbf{47.25}\%  &\underline{47.07\%}\\
    \textbf{CK+}	&30.15\% &31.07\% &36.16\% &39.52\% &40.17\% &39.42\% &41.15\% &40.71\% &39.14\% &40.14\%\\ \bottomrule	
    &&&&&&&&&&\\ \toprule	
    
    \multirow{2}{*}{\pbox{10cm}{\textbf{FER \newline dataset}}}&\multicolumn{10}{c}{\textbf{$\mathbf{x_3}$ value for FI IER dataset}}\\\cmidrule{2-11}
    &0.1&0.2&0.3&0.4&0.5&0.6&0.7&0.8&0.9&1.0\\\midrule
    \textbf{FER13}	&33.24\% &35.92\% &41.14\% &43.43\% &45.55\% &47.71\% &47.67\% &49.96\% &45.37\% &42.46\%\\
    \textbf{FERG}	&26.17\% &28.48\% &30.61\% &32.77\% &36.17\% &39.56\% &41.38\% &41.15\% &40.76\% &39.52\%\\
    \textbf{JAFFE}	&28.43\% &29.85\% &28.71\% &34.18\% &39.92\% &37.57\% &37.13\% &35.29\% &34.03\% &35.57\%\\
    \textbf{CK+}	&35.29\% &38.86\% &39.27\% &42.96\% &44.57\% &45.83\% &51.39\% &\textbf{57.47}\% &\underline{54.82\%} &52.27\% \\\bottomrule	
    &&&&&&&&&&\\ \toprule	
    
    \multirow{2}{*}{\pbox{10cm}{\textbf{FER \newline dataset}}}&\multicolumn{10}{c}{\textbf{$\mathbf{x_4}$ value for EMOTIC IER dataset}}\\\cmidrule{2-11}
    &0.1&0.2&0.3&0.4&0.5&0.6&0.7&0.8&0.9&1.0\\\midrule
    \textbf{FER13}	&32.24\% &40.41\% &41.43\% &43.59\% &44.51\% &46.49\% &46.29\% &45.82\% &46.48\% &\underline{47.17\%}\\
    \textbf{FERG}	&30.39\% &38.34\% &41.51\% &42.53\% &44.84\% &43.84\% &44.48\% &43.58\% &44.58\% &43.61\%\\
    \textbf{JAFFE}	&35.63\% &41.16\% &42.63\% &43.41\% &43.26\% &44.15\% &46.12\% &\textbf{47.62}\% &46.53\% &46.72\%\\
    \textbf{CK+}	&34.13\% &35.93\% &37.49\% &41.15\% &42.47\% &40.48\% &41.92\% &40.57\% &40.59\% &39.94\%\\ \bottomrule	
\end{tabular} 
}
}
\end{table}  

\item 
This approach trains a network for FER and IER simultaneously and minimizes the loss between them during the training. Further, the following networks have been considered while formulating the baseline models and proposed system -- AlexNet (Baseline\#1),  VGG16 (Baseline\#2),  ResNet (Baseline\#3), DeepEmotion \cite{minaee2021deep} (Baseline\#5), and the FER model proposed in section \ref{sec5.1.2} (Proposed Method). Table \ref{tab:da_exp} summarizes their respective performances. The rationale for choosing DeepEmotion is its state-of-the-art performance for FERG, JAFFE and CK+ datasets \cite{malik2021towards}. The individual FER datasets and their combination (denoted as `All') have been used during the implementation. In most cases, the experiments combining all FER datasets have performed better than the individual FER datasets. 

\begin{table}[]
\centering
{\fontsize{10}{12}\selectfont
\caption[]{Results for FER to IER domain adaptation experiments. The notation `F' $\rightarrow$ `I' indicates that the IER task on dataset `I' has been adapted from the FER task on dataset `F.' The best and second-best results are highlighted in bold and underline, respectively.}
\label{tab:da_exp}
\resizebox{1\textwidth}{!}
{
\begin{tabular}{lccccc} 
    \multicolumn{6}{c}{IAPSa dataset}\\\toprule
    & \textbf{\small FER13$\rightarrow$IAPSa} & \textbf{\small FERG$\rightarrow$IAPSa} & \textbf{\small JAFFE$\rightarrow$IAPSa} 
    & \textbf{\small CK+$\rightarrow$IAPSa} & \textbf{\small All$\rightarrow$IAPSa} \\\midrule
    Baseline\#1 & 40.24\% & 40.23\% & 42.43\% & 42.83\% & 42.24\% \\
    Baseline\#2 & 48.27\% & 46.24\% & 45.17\% & 48.87\% & 49.25\% \\
    Baseline\#3 & 47.45\% & 46.53\% & 44.31\% & 48.31\% & 48.43\% \\
    Baseline\#4 & 53.73\% & 51.63\% & 50.29\% & 53.72\% & 50.25\% \\
    Baseline\#5 & 57.16\% & 52.57\% & 53.51\% & 57.46\% & 56.54\% \\
    {Baseline\#6} & 56.62\% & 51.34\% & 52.93\% & 56.72\% & 57.81\% \\
    {Baseline\#7} & 55.34\% & 51.67\% & 51.47\% & \underline{59.13}\% & 58.41\% \\
    {Baseline\#8} & 56.72\% & 52.34\% & 53.24\% & 57.93\% & 58.87\% \\
    \textit{Proposed}   & 57.24\% & 53.13\% & 53.36\% & \textbf{61.86}\% & 58.93\% \\ \bottomrule
    & & & & &\\
    
    \multicolumn{6}{c}{ArtPhoto dataset}\\\toprule
    & \textbf{\small FER13$\rightarrow$ArtPhoto} & \textbf{\small FERG$\rightarrow$ArtPhoto} & \textbf{\small JAFFE$\rightarrow$ArtPhoto} 
    & \textbf{\small CK+$\rightarrow$ArtPhoto} & \textbf{\small All$\rightarrow$ArtPhoto} \\\midrule
    Baseline\#1 & 45.87\% & 47.73\% & 46.62\% & 46.28\% & 46.75\% \\
    Baseline\#2 & 50.35\% & 50.85\% & 51.75\% & 48.75\% & 50.24\% \\
    Baseline\#3 & 52.13\% & 52.37\% & 52.42\% & 51.18\% & 51.34\% \\
    Baseline\#4 & 47.82\% & 51.14\% & 48.68\% & 50.23\% & 42.62\% \\
    Baseline\#5 & 54.16\% & 56.16\% & 57.16\% & 54.75\% & 51.45\% \\
    {Baseline\#6} & 53.34\% & 55.74\% & 58.22\% & 53.28\% & 52.34\% \\
    {Baseline\#7} & 54.56\% & 56.12\% & 60.14\% & 54.62\% & 58.42\% \\
    {Baseline\#8} & 53.82\% & 55.89\% & 59.83\% & 55.34\% & 57.29\% \\
    \textit{Proposed}   & 54.62\% & 56.47\% & \textbf{62.47}\% & 56.34\% & \underline{60.27\%} \\\bottomrule
    & & & & &\\
    
    \multicolumn{6}{c}{FI dataset}\\\toprule
    & \textbf{\small FER13$\rightarrow$FI} & \textbf{\small FERG$\rightarrow$FI} & \textbf{\small JAFFE$\rightarrow$FI} 
    & \textbf{\small CK+$\rightarrow$FI} & \textbf{\small All$\rightarrow$FI} \\\midrule
    Baseline\#1 & 56.37\% & 51.97\% & 52.63\% & 54.55\% & 58.21\% \\
    Baseline\#2 & 59.52\% & 56.56\% & 54.56\% & 62.83\% & 59.74\% \\
    Baseline\#3 & 58.17\% & 55.12\% & 51.07\% & 63.18\% & 63.17\% \\
    Baseline\#4 & 59.28\% & 56.97\% & 53.18\% & 57.86\% & 59.45\% \\
    Baseline\#5 & 60.56\% & 56.64\% & 54.35\% & 63.13\% & 63.12\% \\
    {Baseline\#6} & 59.96\% & 57.62\% & 55.67\% & 64.53\% & 66.72\% \\
    {Baseline\#7} & 60.12\% & 56.94\% & 56.35\% & 65.12\% & 65.86\% \\
    {Baseline\#8} & 60.43\% & 57.12\% & 55.98\% & 64.94\% & 67.12\% \\
    \textit{Proposed}   & 61.53\% & 57.67\% & 56.72\% & \underline{68.17\%} & \textbf{70.78}\% \\\bottomrule
    & & & & &\\
    
    \multicolumn{6}{c}{EMOTIC dataset}\\\toprule
    & \textbf{\small FER13$\rightarrow$EMOTIC} & \textbf{\small FERG$\rightarrow$EMOTIC} & \textbf{\small JAFFE$\rightarrow$EMOTIC} 
    & \textbf{\small CK+$\rightarrow$EMOTIC} & \textbf{\small All$\rightarrow$EMOTIC} \\\midrule
    Baseline\#1 & 40.85\% & 39.83\% & 38.25\% & 40.93\% & 41.17\% \\
    Baseline\#2 & 44.94\% & 43.63\% & 42.68\% & 42.93\% & 45.45\% \\
    Baseline\#3 & 44.71\% & 46.73\% & 41.29\% & 43.64\% & 46.32\% \\
    Baseline\#4 & 46.99\% & 40.83\% & 43.73\% & 50.87\% & 47.53\% \\
    Baseline\#5 & 49.44\% & 50.73\% & 49.18\% & 52.86\% & 54.37\% \\
    {Baseline\#6} & 51.46\% & 52.64\% & 51.64\% & 53.75\% & \underline{58.25\%} \\
    {Baseline\#7} & 51.74\% & 53.64\% & 50.86\% & 55.65\% & 57.94\% \\
    {Baseline\#8} & 50.96\% & 51.86\% & 51.12\% & 55.54\% & 57.83\% \\
    \textit{Proposed}   & 53.16\% & 54.29\% & 53.33\% & 56.36\% & \textbf{59.72}\% \\\bottomrule
\end{tabular} 
}}
\end{table} 

\end{itemize}

Overall, simultaneous training of the FER model proposed in section \ref{sec5.1.2} on FER and IER datasets resulted in better performance. The proposed system implements it.

\subsection{Discussion}
As more pre-trained models and labelled large-scale datasets are available for FER, we have proposed a domain adaptation-based system to use them for IER. Our domain adaptation approach utilizes a feature-based strategy, retraining the same architecture for IER as was developed for FER, coupled with discrepancy loss. This method ensures the effective transfer of learned behaviours from FER to IER, even when applied to small-scale IER datasets. By maintaining the same architecture, we facilitate a seamless domain adaptation process, optimizing the model's performance through a nuanced understanding of both facial and non-facial emotion indicators. The use of discrepancy loss helps in aligning the target domain's distribution with the source, further enhancing the model's efficacy.

Accuracy has been adopted as the primary performance measure for its direct representation of the proportion of true results within the total number of cases examined. This metric has been specifically chosen because it provides a clear and understandable measure of model performance across varied datasets. In the presentation of our results in Table \ref{tab:ier_sota}, we have included only those existing works for which accuracy values are available. While many studies employ different metrics such as mAP or AUC, focusing on accuracy allows for a consistent and straightforward comparison across different approaches.

The proposed interpretability approach, DnCShap, has effectively highlighted important features within images that contribute to emotion recognition. Observations from Fig. \ref{fig:fmapsIER} show that while the feature maps for the `happy' and `anger' classes are more precise, those for the `sad' and `hate' classes appear more scattered. This variance provides valuable insights into how different emotions are processed and recognized by the system, suggesting areas for further fine-tuning. Moreover, extending interpretability to other modalities may yield additional improvements and insights, enhancing the system's overall robustness and applicability.

An intriguing aspect of the results presented in Fig. \ref{fig:cm} is the variability in how similar emotions are recognized across different datasets. This could be attributed to the inherent differences in dataset composition and the contexts within which the images were captured. Each dataset may capture unique emotional expressions influenced by cultural nuances and the specific circumstances of image capture, affecting the generalizability of the results. Further, the dependency of performance on specific datasets, as evidenced in Table \ref{tab:da_exp}, underscores the impact of each dataset's unique characteristics on emotion recognition efficacy. Datasets with artistic images might encapsulate subtler emotional expressions compared to those with more expressive images from controlled settings. Understanding these characteristics is crucial for refining the training and adaptation processes of IER systems.\vspace{.05in}

{\textit{Generalizability Across Datasets}:
Our proposed method has demonstrated robust performance across standard datasets used in facial emotion recognition. However, the scalability and generalizability of the system to other datasets, especially those that might contain more complex or abstract emotional cues, warrant further exploration. The architecture and methodologies employed in our system are designed to be adaptable and flexible, allowing for potential extensions to datasets encompassing a broader range of emotional expressions and cultural variations. For example, the inclusion of domain adaptation techniques such as discrepancy loss ensures that our model can learn domain-invariant features, which is crucial for applications involving diverse emotional datasets. Future work will focus on testing the model's performance on datasets like the Emotionet, which includes a wide array of subtle and complex emotional expressions, often captured in less controlled environments. Additionally, we plan to explore the integration of multimodal data sources, such as audio and textual information, to enrich the emotional context and enhance the model's interpretability and adaptability to real-world scenarios where emotional cues are not solely visual.\vspace{.05in}
}

{\textit{Failure Case Analysis}: 
Despite achieving competitive accuracy across multiple datasets, our model exhibits certain failure cases that warrant further analysis. A key challenge arises in the misclassification of visually similar emotions, such as confusion between fear and surprise or sadness and neutral expressions, particularly in the FER domain. In the IER setting, errors frequently occur in images where emotional cues are highly context-dependent rather than explicitly conveyed through facial expressions. Additionally, domain adaptation limitations contribute to errors when transferring knowledge from FER to IER, especially when the dataset distributions differ significantly. Low-quality images, occlusions, and dataset biases further affect the model's performance, leading to occasional inconsistencies in predictions. To mitigate these issues, potential improvements include incorporating uncertainty-aware learning, improving dataset balance, and leveraging multi-modal cues beyond visual information to enhance emotion recognition reliability. This analysis highlights the importance of refining adaptation strategies and dataset curation to address these failure cases effectively.
}

\section{Conclusion and Future Work}\label{sec:conclusion}	
This paper has introduced a novel feature-based domain adaptation strategy that effectively uses discrepancy loss to align the IER domain with the FER domain, employing the same network architecture for both tasks. Our approach not only demonstrates robust emotion recognition across diverse image contexts but also achieves notable accuracy metrics across multiple datasets. The integration of the DnCShap interpretability method enriches our understanding of key visual features influencing emotion recognition, enhancing transparency and trust in AI-driven systems. Looking ahead, the method’s potential applicability in areas lacking large-scale datasets invites further exploration, particularly in real-world scenarios where emotional contexts are highly variable. Future work should focus on refining the domain adaptation techniques to meet the nuanced demands of these applications and expanding the interpretability framework to include more complex emotional scenarios and diverse datasets. Additionally, exploring the integration of multimodal data, such as audio and textual cues, could provide a more comprehensive understanding of emotional states and further improve the accuracy and robustness of emotion recognition systems. This ongoing research will aim to enhance the model's scalability, adaptability, and efficiency, moving toward more sophisticated and universally applicable emotion recognition systems.\vspace{.2in}

\section*{Declarations}
\noindent \textbf{Acknowledgments}: The majority of this work was conducted at the Machine Vision Lab, Indian Institute of Technology Roorkee, India. The final stages of manuscript preparation and submission were carried out by the primary author while working at the University of Oulu, Finland.\vspace{.2in}

\noindent \textbf{Funding}: This research was supported by Ministry of Human Resource Development (MHRD) INDIA with reference grant number: OH-31-23-200-428.\vspace{.2in}

\noindent \textbf{Competing interests}: Authors have no conflict of interest.\vspace{.2in}

\noindent \textbf{Code availability}: available on request.\vspace{.2in}

\noindent \textbf{Availability of data and material}: available on request.\vspace{.2in}

\noindent \textbf{Authors' contributions}: \textit{Puneet Kumar}: Data Curation, Methodology, Implementation, Experiments, Result Analysis, Writing - original draft \& editing. \textit{Balasubramanian Raman}: Project administration and review.\vspace{.2in} 

\noindent \textbf{Ethics approval}: `Not applicable'.\vspace{.2in}

\noindent \textbf{Consent to participate}: `Not applicable'.\vspace{.2in}

\noindent \textbf{Consent for publication}: `Not applicable'.\vspace{.2in}

\noindent This article does not contain any studies with human participants or animals performed by any of the authors	

\begin{appendices}
\section{}\label{Appendix}
The proposed interpretability DnCShap has been mathematically illustrated in Algorithm \ref{algo:1}, which uses Algorithm \ref{algo:2} for recursive calculations. 

\begin{algorithm}[H]
{
	\textbf{Define} $model$: DNN model,\\ 
	\textbf{Define} $data$: Image pixels, \\ 
	\textbf{Define} $wd, ht$: Image's width \& height,\\ 
	\textbf{Output} $SHAP\_value$.\vspace{.025in}

    \textbf{procedure} \textbf{DnCShap} ($model,\ data,\ wd,\ ht,\ times$)\\
    $data\_b = np.zeros([wd, ht, 3])$\\
    $data\_f = data$\\
    $data\_f = data\_f.reSHAPe(1, wd, ht, 3)$\\
    $data\_b = data\_b.reSHAPe(1, wd, ht, 3)$\vspace{.05in}
    
    $\triangleright$ Perturb by dividing image into two parts, along x or y axis\\
    $data\_1, data\_2 = get\_parts(data,width,height,x)$\vspace{.05in}
    
    $pred = model.predict(data\_f)$\\
    $arg\_{max} = np.argmax(pred)$\\
    $pred\_f = pred[0][arg\_max]$\\
    $pred\_b = model.predict(data\_b)[0][arg\_max]$\\
    $pred\_1 = model.predict(data\_1)[0][arg\_max]$\\
    $pred\_2 = model.predict(data\_2)[0][arg\_max]$\\
    $score\_1 = ((pred\_1-pred\_b) + (pred\_f-pred\_2))/2$\\
    $score\_2 = ((pred\_2-pred\_b) + (pred\_f-pred\_1))/2$\\
    $SHAP\_value = np.zeros([width,height])$\\
    $times = times-1$\vspace{.05in}
    
    \textbf{RecSHAP} ($model, data, wd, ht, times, 1, 0, x\_m, 0, ht, pred\_b+pred\_2,$\\ $pred\_1+pred\_f, score\_1, arg\_max, SHAP\_value$)
    
    \textbf{RecSHAP} ($model, data, wd, ht, times, 1, x\_m, wd, 0, ht, pred\_b+pred\_1,$\\ $pred\_2+pred\_f, score\_2, arg\_max, SHAP\_value$)\\\vspace{.025in}
    return $SHAP\_value$
    \caption{DnCShap}
	\label{algo:1} 
}
\end{algorithm}

\begin{algorithm}[H]
{
	\textbf{Define} $model$: DNN model,\\ 
	\textbf{Define} $data$: Image pixels, \\ 
	\textbf{Define} $wd, ht$: Image's width \& height,\\ 
	\textbf{Define} $pred\_b, pred\_f$: prediction with both/no parts perturbed, \\ 
	\textbf{Define} $x\_s,x\_e$ initial \& final dimensions for width\\
	\textbf{Define} $y\_s,y\_e$ initial \& final dimensions for height\\
	\textbf{Define} $times$: number of parts the image is to be divided, \\ 
	\textbf{Define} $arg\_{max}$: predicted label, \\ 
	\textbf{Define} $score$: SHAP value of the part, \\ 
	\textbf{Define} $value$: activation map.\vspace{.025in}
	
	\textbf{procedure} \textbf{RecSHAP} ($model,\ data,\ wd,\ ht,\ times,\ i, x\_s,\ x\_e,$\\
	$y\_s,\ y\_e,\ pred\_b,\ pred\_f,\ score,\ arg\_{max},\ value$)\vspace{.025in}
	
    if (times == 0) then\\
       \hspace*{1em} $value[x_s:x_e, y_s:y_e]$ = score;\\
    else (if i == 1) then\\
       \hspace*{1em} z = y;\\
    else\\
       \hspace*{1em} z = x;
    
    $\triangleright$ Perturb by dividing image into two parts, along x or y axis\\
    $data\_1, data\_2 = get\_parts(data,\ wd,\ ht,\ z)$\vspace{.05in}

    
    $pred\_1 = model.predict(data\_1)[0][arg\_{max}]$\\
    $pred\_2 = model.predict(data\_2)[0][arg\_{max}]$\vspace{.05in}
    
    $score\_1 = (((pred\_1-pred\_b) + (pred\_f-pred\_2))/2)/2$\\
    $score\_2 =  (((pred\_2-pred\_b) + (pred\_f-pred\_2))/2)/2$\\
    $times = times - 1$\\
    $i = abs(i-1)$\vspace{.05in}
    
    \textbf{RecSHAP} ($model,\ data,\ wd,\ ht,\ times,\ i,\ x\_s,\ x\_e,\ y\_s,\ y\_e$,\\
    $pred\_b+pred\_2,\ pred\_1+pred\_f,\ score\_2,\ arg\_{max},\ value$)
	
    \textbf{RecSHAP} ($model,\ data,\ wd,\ ht,\ times,\ i,\ x\_s,\ x\_e,\ y\_s,\ y\_e$,\\
    $pred\_b+pred\_1,\ pred\_2+pred\_f,\ score\_2,\ arg\_{max},\ value$)
    
	\caption{RecSHAP (recursive procedure)}
	\label{algo:2}
}
\end{algorithm}

\end{appendices}

\bibliography{output} 
\end{document}